\documentclass[journal ]{new-aiaa}
\usepackage[utf8]{inputenc}
\usepackage{textcomp}
\usepackage{booktabs}
\usepackage{graphicx}
\usepackage{amsmath}
\usepackage[version=4]{mhchem}
\usepackage{siunitx}
\usepackage{longtable,tabularx}
\usepackage{algorithm}
\usepackage{algorithmic}
\usepackage{graphicx}      
\usepackage{subcaption}    
\usepackage{float}         

\usepackage{enumitem}
\usepackage{float}
\title{From Sentiment to Actionable Insights: Public Sentiment Analysis of Advanced Air Mobility}

\author{
Esrat Farhana Dulia\textsuperscript{1}\footnote{Graduate Research Assistant, College of Aeronautics and Engineering, edulia@kent.edu.}, 
Amina Dhaher\textsuperscript{1}\footnote{Undergraduate Research Assistant, College of Aeronautics and Engineering, adhaher@kent.edu.}, 
Raiful Hasan\textsuperscript{1}\footnote{Assistant Professor, Department of Computer Science, rhasan7@kent.edu.}, and
Syed Arbab Mohd Shihab\textsuperscript{1}\footnote{Assistant Professor, College of Aeronautics and Engineering, sshihab@kent.edu.}
}

\affil{
\textsuperscript{1}Kent State University, Kent, OH, USA 44242 \\
}

\begin{document}

\maketitle

\begin{abstract}

Advanced Air Mobility (AAM) is an emerging low-altitude transportation system whose successful deployment depends on both technological progress and public acceptance. Public acceptance can influence government support, regulations, noise standards, willingness to fly, and the commercial viability of AAM. Understanding public sentiment is therefore essential for identifying societal barriers and developing effective adoption strategies. This study analyzes 306,009 human-generated texts collected from Reddit and Quora to examine AAM-related public discourse using artificial intelligence models. Seven sentiment-analysis approaches, including lexicon-based, machine-learning, deep-learning, and transformer models, are evaluated to identify the most reliable method for AAM-specific sentiment classification. ModernBERT achieves the highest performance and is used to label the full dataset. Latent Dirichlet Allocation is then applied within each sentiment class to identify underlying topics and examine their temporal evolution from 2008 to 2025. The analysis identifies 20 topics and six major cross-sentiment clusters: workforce and skill development, regulation and compliance, drone technical performance, military and geopolitical applications, safety and operational risks, and noise and disturbance. These findings can help policymakers, industry stakeholders, researchers, and operators develop targeted regulations, safety measures, workforce programs, noise-reduction strategies, and public communication efforts to address concerns and support the responsible deployment of AAM.

\end{abstract}

\section{Introduction}

\subsection{Advanced Air Mobility as a Socio-Technical Air Transport System}

AAM has the potential to provide economic, social, and environmental benefits by improving mobility, increasing transportation accessibility, reducing travel time, and creating new opportunities for passenger and cargo transportation \citep{Dulia2021AAM,DelRosario2021Infrastructure,open_framework_standards,Calhoun2023OpenFramework}. To support the safe and efficient deployment of AAM, federal agencies, industry, and academia have studied various technical and operational challenges, including air traffic management \citep{FAA2020}, concepts of operation \citep{thipphavong2018urban}, market analysis \citep{hasan2018urban}, infrastructure investment planning \citep{dulia2024building,dulia2024negotiate,dulia20263r}, operational planning \citep{Dulia2025LowNoiseUAV,dulia2025dynamic}, and supply chain planning \citep{dulia2024integrated}. These studies have advanced the understanding of AAM technologies and operational requirements. However, AAM is not only an aviation technology system; it is a socio-technical air transportation system that combines electric vertical takeoff and landing aircraft, autonomous flight technologies, digital communication networks, low-altitude traffic management, regulatory frameworks, and public acceptance \citep{nasaAAM2023}. 

Unlike conventional aviation, which mainly operates through established airports and controlled routes, AAM is expected to use distributed vertiports, low-altitude traffic corridors, and frequent operations in urban areas \citep{thipphavong2018urban}. Because AAM operations will occur close to communities, successful deployment will depend not only on aircraft and airspace technologies but also on how users, communities, and regulators perceive and interact with the system \citep{yoo2022risk,lee2023societal,nationalacademies2020aam}. Concerns related to safety, noise, affordability, equity, and trust in automation can influence public willingness to adopt AAM services and affect decisions regarding certification requirements, vertiport locations, operational limitations, and acceptable traffic levels \citep{dallenbach2022far,elias2012influence}. Therefore, public perception and sentiment should be considered as part of the AAM system design and deployment process rather than treated as an independent social factor.

The socio-technical nature of AAM becomes more important when technological progress is considered alongside implementation challenges. Considerable advances have been made in eVTOL design, propulsion, flight control, aeroacoustics, certification, and low-altitude traffic management \citep{xiang2024autonomous,doo2021nasa,aposporis2024review}. These developments have improved the technical feasibility of AAM and established preliminary concepts for high-density urban operations \citep{kopardekar2016unmanned,cohen2021urban}. However, experience with high-speed rail, automated transit, and connected autonomous vehicles shows that technical readiness does not guarantee timely implementation when regulatory issues and public concerns remain unresolved \citep{albalate2012high,furman2014automated,ahmed2022technology}. For AAM, this means that aircraft, digital infrastructure, regulation, and public acceptance should be examined together. While engineering research has studied the physical and digital components in detail, the societal component remains less developed, creating a need for methods that can incorporate public opinion into AAM planning and decision-making.

\subsection{Assessing Public Sentiment through Online Platforms}

Assessing public sentiment through online platforms provides a way to examine how communities discuss transportation systems outside structured surveys. Transportation projects such as highway expansions, rail systems, and airport developments have faced delays, redesign, or public opposition because of concerns about noise, environmental impacts, safety, accessibility, and equity \citep{nguyen2024significant,chwilkowska2020sources}. These concerns are often studied through surveys, interviews, and questionnaires \citep{chen2021using,nakshi2025using}. Although such methods provide organized and statistically analyzable information, they are often limited by small samples, fixed response options, and one-time data collection. They may also overlook concerns that emerge through local discussions, public meetings, or changing social conditions \citep{nakshi2025using,dougald2022analysis}.

Online platforms extend this assessment by allowing users to discuss issues in their own words, respond to others, and revisit topics as new events occur. Reddit, Quora, and Twitter have been used to study public reactions to transportation services, safety concerns, and policy changes \citep{melton2021public,bhaduri2026shifting,bakalos2020public}. Their content is less structured than survey data, but it can reveal concerns and expectations that researchers may not identify in advance. This is especially useful for AAM because its regulations, infrastructure, and operating practices are still evolving. Analyzing online discussions can therefore help identify how the public interprets safety, automation, noise, regulation, workforce effects, and other issues that may influence deployment.

\subsection{Using AI to Analyze Public Sentiment}

Using artificial intelligence to analyze public sentiment makes it possible to process online discussions at a scale that manual review cannot support. AI-based methods can classify sentiment, identify recurring topics, and track how public opinion changes over time \citep{mao2024sentiment}. These methods have been applied in healthcare, finance, energy, environmental policy, and transportation to examine public concerns and responses to policy or service changes \citep{gao2020public,nguyen2015sentiment,el2021novel,jeong2023public,saha2025public}. Their use shows how large collections of unstructured text can be converted into organized evidence about public attitudes and recurring themes.

For AAM, AI-based sentiment analysis can reveal patterns that may not be captured through surveys alone. It can compare positive and negative discussions, identify domain-specific concerns, and examine how topics change over time across different online communities. However, this requires domain-specific training because AAM discussions contain technical terminology, regulatory references, and context-dependent expressions that general sentiment models may interpret incorrectly. The present study addresses this need by comparing several sentiment-analysis methods and applying the best-performing model to a large Reddit and Quora dataset.

\subsection{Contributions of This Study}

The contributions of this study address the need for a larger, domain-specific, and methodologically broader analysis of public sentiment toward AAM:

\begin{itemize}[leftmargin=0pt,labelindent=0pt,itemindent=0pt,
                labelsep=0.6em,itemsep=0pt,topsep=0pt,
                parsep=0pt,partopsep=0pt]

\item The study develops a domain-specific dataset containing 306,009 AAM-related texts collected from Reddit and Quora.

\item The study compares seven sentiment-classification models representing rule-based, classical machine learning, recurrent deep learning, hybrid deep learning, and transformer approaches: VADER \citep{Hutto2014VADER}, Naive Bayes \citep{noori2021classification}, BiLSTM with attention \citep{Schuster1997BiRNN}, Text CNN--BiLSTM \citep{Li2024research}, RoBERTa \citep{Liu2019RoBERTa}, DeBERTa \citep{He2021DeBERTa}, and ModernBERT \citep{warner2025smarter}.

\item The study creates a labeled dataset of 5,000 texts using human and AI-assisted annotation and then uses the best-performing model, ModernBERT, to classify the full dataset.

\item The study integrates sentiment classification, Latent Dirichlet Allocation (LDA) topic modeling, cross-sentiment comparison, and temporal analysis to identify 20 topics and examine their development from 2008 to 2025.

\item The study groups these topics into six broader areas: regulation and compliance, safety and operational risks, technical performance, military and geopolitical issues, workforce and career development, and noise and disturbance.

\item The study uses these findings to discuss planning and policy strategies intended to improve public trust, address adoption barriers, and support socially acceptable AAM deployment.

\end{itemize}

\section{Literature Review}
\label{sec_Lit_rev}

\subsection{Public Sentiment and Acceptance of Advanced Air Mobility}

Public acceptance is a major requirement for the successful introduction of emerging transportation technologies. Research on autonomous vehicles, unmanned aerial systems, and smart mobility shows that adoption is strongly influenced by perceived safety, trust in automation, environmental concerns, regulatory clarity, usefulness, anxiety, and perceived loss of control \citep{Bansal2016AV,Madigan2017ARTS,Nordhoff2019AV}. Similar factors have been reported in Advanced Air Mobility (AAM) research, where safety, noise, affordability, travel-time savings, operational transparency, and trust in autonomous systems affect willingness to use air taxi and related services \citep{AlHaddad2020UAM,Chancey2020UAM}. For example, \citet{AlHaddad2020UAM} found that perceived safety, ease of use, and travel-time benefits influence AAM adoption, while \citet{Chancey2020UAM} showed that trust in automation affects acceptance of remotely operated air taxis. More recent work, including \citet{guan2025understanding}, continues to rely mainly on Likert-scale surveys and stated-preference methods.

Although these methods provide structured and statistically analyzable results, they limit respondents to predefined questions and response categories. As a result, they may not capture spontaneous concerns, changing narratives, or emotionally detailed opinions that appear in open online discussions. This limitation is important for AAM because public attitudes may shift quickly following regulatory changes, safety incidents, technology demonstrations, or media coverage. To examine these broader discussions, \citet{Bhaduri2026ShiftingSkies} applied Natural Language Processing (NLP) to 15,442 AAM-related tweets and used 1,365 annotated tweets for model training. Their support vector machine and BERT models achieved accuracies of 0.725 and 0.751, respectively. While this study demonstrated the value of NLP for AAM sentiment analysis, the moderate classification performance and limited annotated sample show the need for larger datasets, richer discussion platforms, and more recent language models. These needs motivate a broader review of sentiment-analysis methods.

\subsection{Natural Language Processing Approaches for Sentiment Analysis}
\label{sec_Lit_rev_NLP_models}

Natural Language Processing is a branch of artificial intelligence that enables computers to analyze and interpret human language. One of its main applications is sentiment analysis, which classifies opinions or attitudes expressed in text. Existing methods include lexicon-based systems, classical machine learning, unsupervised and semi-supervised methods, recurrent deep learning models, and transformer architectures. These approaches differ in their need for annotated data, use of manual features, computational requirements, and ability to represent context.

\begin{table}[h]
\centering
\caption{Comparison of major sentiment analysis approaches.}
\label{tab:sentiment_model_comparison}
\begin{tabular}{p{2.5cm}p{4.3cm}p{7.2cm}}
\hline
\textbf{Approach} & \textbf{Representative Models} & \textbf{Key Characteristics} \\
\hline
Lexicon-based &
VADER, SentiWordNet &
Use predefined sentiment dictionaries; efficient and interpretable but limited in capturing context and domain-specific language. \\

Machine learning &
Naive Bayes, SVM, logistic regression &
Use engineered features such as TF-IDF and n-grams; computationally efficient but limited in modeling word order and contextual meaning. \\

Unsupervised and semi-supervised &
PMI, LDA, JST, LSA, NMF, graph propagation &
Extract sentiment or latent structures with limited annotated data, but often provide less precise and less interpretable classifications. \\

Recurrent deep learning &
RNN, LSTM, BiLSTM, CNN-BiLSTM &
Capture sequential and local contextual patterns with limited feature engineering, but require sequential computation and substantial training data. \\

Transformers &
BERT, RoBERTa, DeBERTa, ModernBERT &
Use self-attention to capture long-range context and domain-specific meaning; generally achieve strong classification performance but require greater computational resources. \\
\hline
\end{tabular}
\end{table}

\subsubsection{Lexicon-Based Methods}

Lexicon-based methods assign sentiment scores to words and combine them to estimate document-level polarity \citep{Taboada2011Lexicon,Liu2012SentimentAnalysis}. VADER improved this approach by accounting for punctuation, capitalization, negation, and intensity modifiers, making it suitable for short and informal text \citep{Hutto2014VADER}. These methods are transparent, computationally efficient, and useful when annotated data are limited \citep{van2025advantages,cero2024lexicon}. However, they depend on static and often domain-independent vocabularies, which makes it difficult to interpret technical terms, sarcasm, implicit sentiment, and context-dependent expressions \citep{deng2017adapting,hamilton2016inducing,sadia2018overview}. In AAM discussions, terms such as ``eVTOL,'' ``vertiport,'' and ``autonomy'' may carry meanings that general sentiment dictionaries do not capture correctly. Therefore, lexicon-based methods are valuable as baselines, but they are less suitable for detailed domain-specific sentiment classification.

\subsubsection{Classical Machine Learning Methods}

Classical supervised models, including Naive Bayes, support vector machines, and logistic regression, learn relationships between annotated sentiment labels and textual features such as bag-of-words, n-grams, and TF-IDF values \citep{pang2002thumbs,Mullen2004SVM,Go2009Twitter}. These models are relatively efficient, interpretable, and effective when moderate-sized labeled datasets are available. They have also been used in aviation research to classify Aviation Safety Reporting System narratives and identify hazards \citep{Oztekin2013ASRS,Li2016AviationText}. However, because they rely on sparse feature representations, they do not adequately capture word order, long-range context, negation, or technical expressions and often require substantial feature engineering.

When labeled data are limited, unsupervised and semi-supervised methods can identify patterns directly from unannotated text. These methods include pointwise mutual information, clustering, topic modeling, latent semantic analysis, non-negative matrix factorization, and graph-based propagation \citep{Turney2002Thumbs,Hatzivassiloglou1997Adjectives,Blei2003LDA,Lin2009JST,Deerwester1990LSA,Lee1999NMF}. They are useful for exploratory analysis and large unannotated corpora, but their sentiment categories may not match human interpretations, and their results can be sensitive to seed words, initialization, and corpus composition.

\subsubsection{Deep Learning Methods}

Deep learning models reduce the need for manual feature engineering by learning text representations directly from data \citep{Bengio2009DeepLearning,LeCun2015DeepLearning}. Recurrent neural networks model sequential relationships by passing information across tokens, but standard RNNs are affected by vanishing and exploding gradients \citep{Elman1990RNN,Bengio1994Vanishing}. LSTM networks address this problem through gated memory cells that preserve information over longer sequences \citep{Gers2000LSTMImproved}. BiLSTM models process text in both forward and backward directions, allowing each word to be interpreted using both earlier and later context \citep{Schuster1997BiRNN}. Attention mechanisms further improve these models by assigning greater weight to words that contribute more strongly to sentiment \citep{Bahdanau2015Attention,Yang2016HAN}.

Hybrid CNN-BiLSTM architectures combine local phrase detection with sequential modeling and have achieved stronger performance than standalone CNN or LSTM models in several sentiment tasks \citep{Kim2014CNNText,Wang2016Combination,Zhou2018sentiment,Li2024research}. Despite these improvements, recurrent models process text sequentially, which limits parallel computation and increases training time. Their performance also depends strongly on the size and quality of the annotated dataset.

\subsubsection{Transformer-Based Methods}

Transformer models have become the leading approach to sentiment analysis because self-attention allows them to capture long-range contextual relationships while processing tokens in parallel. BERT introduced bidirectional contextual representations and substantially improved text-classification performance \citep{Devlin2019BERT}. RoBERTa later improved BERT through larger training corpora and optimized pretraining \citep{Liu2019RoBERTa}, while DeBERTa introduced disentangled attention to represent content and position more effectively \citep{He2021DeBERTa}. ModernBERT further extends this development through longer context windows and efficiency improvements for large-scale applications \citep{warner2025smarter}.

\begin{table}[h]
\centering
\caption{Key characteristics of selected transformer models.}
\label{tab:transformer_model_comparison}
\begin{tabular}{p{2.2cm}p{5cm}p{7cm}}
\hline
\textbf{Model} & \textbf{Architecture} & \textbf{Main Improvement} \\
\hline
BERT &
Bidirectional transformer with masked language modeling &
Introduced widely used bidirectional contextual representations. \\

RoBERTa &
Optimized BERT architecture &
Uses larger corpora and improved pretraining without next-sentence prediction. \\

DeBERTa &
Disentangled content and position attention &
Improves contextual and positional representation. \\

ModernBERT &
Modernized encoder with long context and efficient attention &
Improves scalability, speed, and contextual modeling for large text datasets. \\
\hline
\end{tabular}
\end{table}

Transformer models generally outperform classical and recurrent approaches because they learn richer contextual representations and can be fine-tuned for specific domains \citep{Minaee2021DeepSentiment,Sun2019BERTSentiment}. Social-media-specific models such as BERTweet further improve performance on informal language, abbreviations, and platform-specific expressions \citep{Nguyen2020BERTweet,Barbieri2020TweetEval}. These strengths make transformers suitable for AAM discourse, where sentiment often depends on technical vocabulary, policy context, and relationships among multiple words. Their main disadvantages are high computational cost, memory requirements, possible overfitting on small labeled datasets, and limited interpretability. Even with these limitations, transformer models remain the strongest candidates for large-scale, domain-specific sentiment classification.

\subsection{Research Gaps in Understanding Public Sentiment Toward AAM}

In the reviewed literature, NLP-based analyses of online discourse have gained increasing attention due to the rapid growth of large-scale data from social networks and online forums. These platforms provide rich sources of spontaneous public discussion and have been widely used for sentiment analysis across various domains. However, such approaches remain relatively limited in the context of AAM. Most existing studies examining public perception of AAM rely primarily on survey-based methods that use predefined constructs and closed-ended responses. Although surveys provide structured and statistically analyzable insights, they often fail to capture spontaneous public discourse, evolving narratives, and emotionally nuanced opinions that frequently emerge in open online discussions. Given the rapid expansion of online discussion platforms, there is a growing need for research that uses NLP techniques to analyze large-scale user-generated content and better understand public sentiment toward emerging mobility technologies.

To the best of our knowledge, the only study applying NLP to AAM-related public discourse is the work of \citep{Bhaduri2026ShiftingSkies}, which employed both a classical machine learning model (SVM) and a transformer-based model (BERT) to analyze Twitter discussions. Their dataset consisted of 15,442 tweets, with 1,365 annotated tweets used for model fine-tuning. Although the study demonstrates the potential of NLP approaches for analyzing public discourse about AAM, the relatively small annotated dataset limits the amount of supervised data available for effective transformer training. Consequently, the reported model performance remained moderate, with SVM achieving an accuracy of 0.725 and BERT achieving 0.751. Expanding the annotated dataset and the overall corpus for domain-specific text is therefore critical. Larger datasets enable more effective transformer fine-tuning, leading to higher classification accuracy, improved capture of domain-specific terminology, and more reliable labeling for subsequent topic modeling. This produces higher-quality insights into public sentiment and topics, providing a reliable understanding of public acceptance in AAM.

While the Twitter-based approach adopted by \citep{Bhaduri2026ShiftingSkies} provides valuable initial insights into public discourse on AAM, incorporating additional online discussion platforms can further broaden and complement the overall understanding of public sentiment. In particular, platforms such as Reddit and Quora support more sustained, contextualized, and interactive exchanges through threaded discussions and detailed responses. Unlike Twitter, which is primarily characterized by short and fragmented posts, Reddit and Quora enable users to elaborate on viewpoints, justify opinions, and engage in follow-up dialogue \citep{CurrentTrends2026FragmentedSocial}, thereby providing richer semantic content for sentiment and topic analysis. Integrating these platforms also enhances demographic representation beyond Twitter’s user base. In 2025, Twitter reported approximately 557 million monthly active users, with the dominant age group being 25--34 years (37.5\%) \citep{Jaimes2025TwitterStats}. Reddit, with over 1 billion monthly active users in 2025, is particularly popular among younger audiences, with users aged 18--29 accounting for approximately 44\% of the platform’s user base \citep{Connell2026RedditStats}, making it especially useful for capturing early-adopter perspectives on AAM. In contrast, Quora reported more than 300 million monthly active users in 2025 and maintains a comparatively balanced and analytically oriented audience, with the largest age groups being 25--34 years (28.36\%) and 18--24 years (25.65\%) \citep{Shewale2026QuoraStats}.

These observations highlight several important gaps in the current literature: (i) limited application of NLP methods for analyzing public sentiment toward AAM using large-scale online discourse, (ii) relatively small datasets, which constrain the effective training and fine-tuning of modern transformer-based models, (iii) limited exploration of newer transformer architectures, with most studies relying on classical machine learning models or earlier transformer-based approaches such as BERT, (iv) reliance on Twitter data alone, which limits viewpoint diversity and overlooks richer, more varied discussions on platforms such as Reddit and Quora, resulting in an incomplete picture of public sentiment toward AAM, and (v) limited focus on translating findings into actionable strategies that can inform policy design and address public concerns effectively. To address these gaps, we conducted this study, which is presented in the following section. To address these research gaps and provide a more comprehensive understanding of public perceptions toward AAM, we conducted the study presented in the following sections.

\section{Methodology}

\begin{figure}[tb!] 
    \centering
    \includegraphics[width=16cm,height=8in]{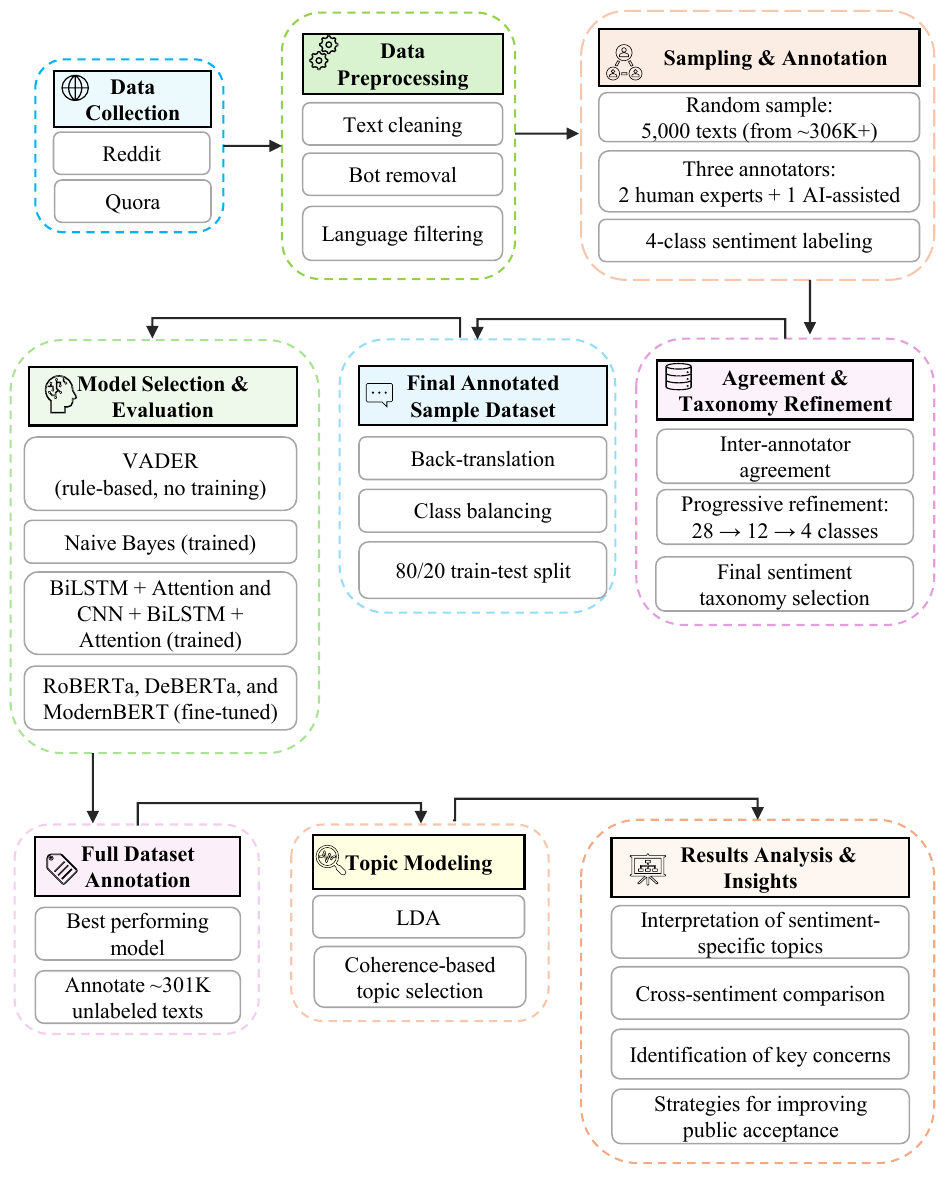}
    \caption{Overview of the methodology for AAM sentiment analysis.} \label{flowchart}
\end{figure}

This section describes the methodology used to analyze public sentiment and topics related to AAM, as shown in Figure \ref{flowchart}. The process begins with data collection from online discussion platforms, followed by preprocessing to clean and filter the data. A subset of the data is then sampled and annotated by three annotators (two human experts and one AI-assisted annotator), and inter-annotator agreement is used to refine the sentiment classes. The final annotated sample dataset is enhanced using back-translation and class balancing to reduce class imbalance and ensure a more equal distribution across sentiment classes. This sample dataset is split into an 80/20 train-test set and used to train and evaluate multiple sentiment classification models to identify the best-performing model for AAM sentiment analysis. Seven models are evaluated: VADER (rule-based, no training), Naive Bayes, BiLSTM with attention, and CNN + BiLSTM with attention (trained on the training set), and RoBERTa, DeBERTa, and ModernBERT (fine-tuned on the same data). The best-performing model is selected on the basis of test accuracy and is then used to annotate the remaining unlabeled data set. Finally, LDA is applied to extract topics within each sentiment class, and the results are analyzed to interpret sentiment-specific topics, compare cross-sentiment patterns, identify key concerns and opportunities, and derive actionable strategies for improving public acceptance of AAM. The individual steps of this pipeline are described in detail in the following subsections.

\subsection{Data Collection}

The first step of our methodology is data collection. We collected textual posts and comments, along with their timestamps and net vote scores (upvotes minus downvotes), from two major online platforms, Reddit\footnote{https://www.reddit.com} and Quora\footnote{https://www.quora.com}, to analyze how people publicly discuss and express their opinions and sentiments  about AAM. Both Reddit and Quora are online platforms where users share perspectives, exchange ideas, and participate in discussions across a wide range of topics. Reddit is a community-driven discussion platform organized into topic-specific subreddits, where users create posts and comment on threads, allowing large-scale public discourse on diverse topics \citep{Reddit2026}. Like Reddit, Quora is also an online platform; however, it focuses mainly on user-generated questions and corresponding answers, often encouraging more detailed explanations and opinions-based responses \citep{Quora2026}.


Our raw dataset consisted of 587,473 texts, including 441,976 Reddit posts and comments and 145,497 Quora questions and answers, which are conceptually similar to Reddit posts and comments. Reddit saw larger data collection and discussion regarding AAM than Quora due to Reddit's larger user base. Reddit sees about 400 million unique active users weekly, while Quora sees about 400 million unique visitors monthly \citep{reddit_press2025, quora_business2025}. In this study, the term “text” refers to a single post, comment, question, or answer, which can contain one or multiple sentences and is treated as one unit for the annotation of sentiments.

\subsubsection{Reddit}
For Reddit data collection, we first reviewed the official Reddit API documentation.
A dedicated Reddit account was created specifically for this project, and an application was registered through Reddit’s developer portal to obtain the required OAuth credentials (client ID, client secret, and user agent). Using these credentials, we developed a Python-based data collection script with the Python Reddit API Wrapper (PRAW) \citep{praw_docs}. 
Reddit’s listing API endpoints, which return collections of posts (e.g., subreddit submissions or user posts), limit each API request for a given listing stream to a maximum of approximately 1,000 retrievable posts. To ensure comprehensive data collection, the script was designed to operate in multiple batches in parallel. 
We adhered to Reddit’s OAuth rate limit of approximately 600 requests per 600 seconds (about one request per second).
To avoid duplicate records, each post was tracked using the unique Reddit identifier, which combines the type of content and the post ID. After completing each batch of data collection, the script stored the identifier of the last post processed. Subsequent scraping sessions resumed from that point, allowing incremental data collection without duplicating previously retrieved post. All posts and their associated comments were saved in structured CSV files for further preprocessing and sentiment annotation.

To identify and filter relevant posts and comments, we constructed a keyword list related to AAM. The keyword list included terms such as ``Advanced Air Mobility,'' ``AAM,'' ``Urban Air Mobility,'' ``UAM,'' ``Unmanned Aerial Vehicle,'' ``UAV,'' ``Unmanned Aircraft System,'' ``UAS,'' ``eVTOL,'' ``Electric Vertical Takeoff and Landing,'' ``UAS Traffic Management,'' ``Vertiport,'' ``Air Taxi,'' ``Flying Car,'' ``Drone,'' ``Quadcopter,'' ``Autonomous Aircraft,'' ``Low-Altitude Airspace,'' ``Electric Aircraft,'' ``Battery-Powered Aircraft,'' ``Vertical Lift Aircraft,'' and ``On-Demand Aviation,'' along with related variations. Here,  drones refers to small Unmanned Aircraft System (sUAS) or small Unmanned Aerial Vehicles (sUAVs). These keywords were applied using case-insensitive exact-phrase searches to retrieve posts containing relevant terminology. Data were collected from multiple subreddits (i.e., topic-specific discussion communities on Reddit) relevant to AAM, including drones, technology, futurology, multicopter, aviation, and 53 additional related communities. By searching for these keywords in these subreddits, we ensured that the dataset captured a wide range of public conversations related to AAM.

\subsubsection{Quora}

For Quora data collection, we used a keyword list identical to that used to retrieve the Reddit posts to gather relevant "questions" posted on Quora. Using Selenium, a Python script was written which would identify any question that contained any one of the keywords and phrases from the keyword list. The script saved the URL's of these questions into a TXT. Using the saved URL's, the script would then record their corresponding answers and save the text into a structured CSV file for further preprocessing. 

\subsection{Data Preprocessing}

To ensure that the collected data were suitable, readable, and understandable for sentiment annotation and model training, we carried out the following steps. (i) First, we corrected mojibake artifacts, which are garbled or misencoded characters resulting from incorrect interpretation of text encoding (e.g., UTF‑8 interpreted as ISO‑8859‑1), and expanded contractions to their canonical forms. For example, the corrupted string ``Their Ã¢â‚¬Å“productÃ¢â‚¬Â is to harvest funds'' was corrected to ``Their product is to harvest funds.'' This step was essential to prevent misinterpreted or invalid tokens that could negatively affect both human annotation and model training. (ii) Next, we removed texts authored by bot-like accounts by filtering usernames containing common automation patterns, such as \texttt{bot}, \texttt{automoderator}, or \texttt{spamaccount}. (iii) Promotional and marketing-related content were excluded using a combination of keyword (e.g., ``buy now,'' ``discount,'' ``promoted'') and pattern matching for URLs or domain-like strings (e.g., endings such as \texttt{.com} or \texttt{.net}). (iv) Posts and comments posted by users annotated \texttt{[deleted]} or \texttt{[removed]} were also excluded, as the original content or author information was no longer available. (v) We further removed texts containing \texttt{!remindme} commands, which are user-generated bot-triggering instructions commonly used on Reddit to request reminder notifications for a post or comment after a specified time interval. These commands are often posted by users (and sometimes bots) to revisit content later or to increase comment engagement, and do not contain meaningful sentiment information. Extremely short greeting-like utterances (e.g., hi,'' hello,'' ok,'' hey,'' yo,'' hmm'') were also excluded, as they do not reliably convey emotional content. (vi) The whitespace was normalized by trimming the leading and trailing spaces and collapsing multiple consecutive spaces into a single space. (vii) Language detection was then applied to retain only English-language texts. (viii) Finally, duplicate texts were removed based on exact text matches, and any texts that were empty or contained only whitespace after preprocessing with above steps were discarded. The preprocessed dataset included 12,459 Quora questions and answers and 293,550 Reddit posts and comments, resulting in a total of 306,009 English-language texts.

\subsection{Sampling and Annotation}

From the cleaned dataset, we randomly sampled 5,000 texts for manual human annotation to create an AAM-specific data set to fine-tune the pretrained AI language models. Although these models are already trained on large, general-purpose datasets, they are not optimized to analyze the technical language, jargon, context, and emotional expressions commonly found in AAM-related texts. Therefore, an annotated sample from our own dataset was required to adapt the models to this domain and to evaluate which model performs best in labeling sentiments before applying it to the entire dataset.

Each sampled text was independently annotated by two human annotators with expertise in the AAM domain and by a Large Language Model (LLM) accessed through the OpenAI API \citep{openai_api}. Human annotators provide an expert understanding of domain-specific terminology and context, which is critical to correctly interpreting emotional content in technical or policy-related discussions. On the other hand, the AI annotator enables a systematic comparison between human judgments and its own predictions. Using both human and AI annotations strengthens the credibility and consistency of the annotated dataset. All annotators, both humans and AI, were instructed to assign exactly one sentiment label per text, based on the dominant emotional signal expressed in the text.

\subsubsection{Inter-Annotator Agreement Metric}





Before using the annotated data for model fine-tuning, we assessed the consistency of sentiment labels across annotators to ensure the reliability of the dataset. Measuring inter-annotator agreement allows us to verify that sentiment classes are interpreted consistently, which is critical for producing high-quality training data. High agreement indicates that the classes are well-defined and understood by annotators. Reliable annotations, in turn, improve the model’s ability to learn accurate patterns, resulting in better sentiment classification and more trustworthy insights into public perceptions of AAM.

Since the annotation involved more than two annotators and sentiment classes, we used Fleiss’s Kappa to quantify the agreement among the annotators \citep{artstein2017inter, moons2025measuring}. This metric assesses the overall consistency among multiple annotators who independently assign labels to the same set of items and is commonly applied in multi-annotator labeling scenarios \citep{fleiss1971measuring, artstein_inter-coder_2008}. Conceptually, Fleiss’s Kappa is calculated from an item-level rating matrix in which each row corresponds to a text and each column records how many annotators assigned that text to a particular sentiment class. Fleiss’s Kappa is defined as

\begin{equation}
\kappa = \frac{\bar{P} - \bar{P}_e}{1 - \bar{P}_e}
\end{equation} 

where $\bar{P}$ represents the mean observed agreement across all items, calculated from the concentration of annotator votes within each class. The expected agreement by chance, $\bar{P}_e$, is derived from the marginal distribution of class labels in all annotations and reflects how often agreement would occur if the annotators assigned labels randomly while preserving the overall class frequencies.

In addition to evaluating how the annotators annotated the texts consistently, we used Fleiss’s Kappa to guide the selection of an appropriate number of sentiment classes. We evaluated different class configurations and selected the one that yielded the highest Kappa value, indicating stronger annotator consistency and a more reliable annotated dataset for model fine-tuning. We also examined how agreement varied across individual sentiment classes to identify classes that were easier or more difficult for annotators to distinguish.

\subsubsection{Progressive Sentiment Classification and Agreement Evaluation}


To determine which sentiment labels are the most appropriate for analyzing public perceptions of AAM, we followed a systematic approach. We initially adopted the 28 fine-grained sentiment labels from the Google GoEmotions taxonomy \citep{demszky-2020-goemotions} as a starting point for our analysis. GoEmotions is a human-annotated dataset of 58,000 Reddit comments collected from popular English-language subreddits that cover a wide range of emotional expressions in online discussions. The 28 labels include \emph{admiration, approval, amusement, anger, annoyance, caring, confusion, curiosity, desire, disappointment, disapproval, disgust, embarrassment, excitement, fear, gratitude, grief, joy, love, nervousness, optimism, pride, realization, relief, remorse, sadness, surprise,} and \emph{neutral}. Using these labels, we measured inter-annotator agreement, which yielded a low Fleiss’ Kappa score of 0.27, indicating only fair consistency among annotators (see Table~\ref{tab:kappa_progression}). This initial set of 28 labels served as our base for further refinement and selection of the most suitable sentiment classes for the AAM context.

\begin{table}[h]
\centering
\caption{Inter-annotator agreement (Fleiss' Kappa) for different sentiment taxonomies}
\label{tab:kappa_progression}
\begin{tabular}{l c l}
\hline
\textbf{Sentiment Taxonomy} & \textbf{Fleiss' Kappa ($\kappa$)} & \textbf{Agreement Level} \\
\hline
28 fine-grained labels & 0.27 & Fair agreement \\
12 broader classes & 0.48 & Moderate agreement \\
4 high-level classes & 0.61 & Moderate to strong agreement \\
\hline
\end{tabular}
\end{table}

To improve annotation consistency, we iteratively merged similar sentiments into broader and clearer classes and recalculated Fleiss’ Kappa after each step. In the first step, the original 28 sentiment labels were merged into 12 broader classes. The \emph{Anger} class included anger, annoyance, and disapproval, while \emph{Disgust} included only disgust. \emph{Fear} combined fear and nervousness, and \emph{Joy} included approval, amusement, excitement, gratitude, joy, love, optimism, pride, relief, and admiration. The \emph{Sadness} class contained disappointment, embarrassment, grief, remorse, and sadness. The remaining classes represented individual sentiments : \emph{Confusion}, \emph{Curiosity}, \emph{Realization}, \emph{Desire}, \emph{Surprise}, \emph{Neutral}, and \emph{Care}. This 12-class taxonomy reduced confusion between closely related sentiments  while preserving meaningful distinctions. As a result, inter-annotator agreement increased to $\kappa = 0.48$, indicating moderate agreement (see Table~\ref{tab:kappa_progression}). Although this represented an improvement, a Fleiss’ Kappa greater than 0.5 is generally not considered sufficient for high‑quality training data \citep{park2022agreement}. Therefore, we aimed for stronger agreement among annotators, since higher Kappa values indicate more consistent labeling, reduced noise in the dataset, and better model performance. This motivated further refinement of the sentiment classes into four high‑level classes.

We further simplified the taxonomy into four high-level sentiment classes: \emph{Positive}, \emph{Negative}, \emph{Neutral}, and \emph{Others}. \emph{Positive} sentiments  included joy, gratitude, love, optimism, and related states, while \emph{Negative} sentiments included anger, fear, sadness, and similar sentiments. \emph{Neutral} captured texts without a clear emotional tone (e.g., “Advanced air mobility pilot programs include testing of air taxi operations, cargo transport, and emergency response applications. They involve collaboration between industry stakeholders and regulatory agencies to assess performance under different use cases.”), whereas \emph{Others} included sentiments of confusion, curiosity, or surprise (e.g., “How will air traffic control systems handle the integration of eVTOL aircraft in already congested urban airspace?”). Under this four-class taxonomy, agreement increased further, with Fleiss’ Kappa rising to $\kappa = 0.61$, corresponding to moderate to strong agreement (see Table~\ref{tab:kappa_progression}). Since this level of agreement ($\kappa = 0.61$) is considered sufficiently strong to produce high-quality training data \citep{marini2021staging}, no further refinement of the sentiment classes was performed.

The progression from 28 to 12 and finally to 4 sentiment classes indicates that although fine-grained sentiment labels are useful for capturing subtle emotional nuances, they can lead to disagreement because closely related sentiments are difficult to distinguish consistently. For instance, consider an AAM-related statement such as: “The rapid expansion of urban air mobility services has left many residents feeling a deep sense of loss about the quiet character of their neighborhoods.” One annotator might label this as \emph{Disappointment} and another as \emph{Sadness}. Although these labels point to a very similar emotional reaction, treating them as distinct categories lowers inter-annotator agreement. The disagreement does not arise because annotators interpret the text differently in spirit, but because the taxonomy forces them to choose between highly overlapping options. When such sentiments are classified into broader classes, the agreement among the annotators improves. Under the 12-class taxonomy, sentiments such as \emph{Disappointment} and \emph{Sadness} are combined, which allows annotators to align more easily on the overall emotional meaning. This effect becomes even stronger in the final 4-class taxonomy, where these sentiments are included within a single negative class. A similar insight is observed for positive sentiments, where labels such as \emph{Optimism}, \emph{Relief}, and \emph{Approval} are captured more consistently when classified under a single positive class. 

Our progressive classification analysis shows that broader sentiment classes yield higher inter-annotator agreement and are more suitable for AAM-related texts. Although both GoEmotions and our dataset are drawn from Reddit, the context of the discussion is very different. GoEmotions was built on diverse and everyday conversations in which sentiments are expressed openly and vividly. In contrast, AAM discussions are typically technical, centered on regulation, safety, infrastructure, and policy, with emotional cues embedded within factual or analytical statements. Because of this difference in tone and expression, applying all 28 fine grained categories can create unnecessary complexity and increase disagreement among annotators. Classifying sentiments into broader categories, such as 12 or 4, better fits the structured nature of AAM discourse and leads to more stable annotations and more reliable model performance in large scale sentiment analysis.


\subsubsection{Final Annotated Sample Dataset}

After applying the four-class sentiment taxonomy and the following agreement-based retention rules, we created a high-confidence annotated dataset. A text was retained if all three annotators agreed, if the two human annotators agreed, or if the two human annotators disagreed but GPT-5 matched one of the human labels. Texts in which all three annotators assigned different labels were discarded. 
Using these rules, 4,550 out of 5,000 annotated texts (about 91\%) were retained, while 450 texts were removed and returned to the main unannotated dataset. 
Among the retained texts, 2,163 had full agreement among all three annotators. The remaining 2,387 texts did not have full agreement among all three annotators. Within these 2,387 texts, 717 had agreement between the two human annotators while GPT-5 assigned a different label. In the remaining 1,670 texts, the two human annotators disagreed, but GPT-5 matched one of the human labels, which was then used as the final label. Each retained text was assigned a single final sentiment label for model training and annotation of the main dataset. As 4,550 texts were annotated using human and OpenAI annotators, a total of 301,459 texts remained in the full dataset as unannotated texts.

\begin{figure}[tb!]
\centering
\begin{subfigure}{\textwidth}
  \centering
  \includegraphics[width=0.7\textwidth]{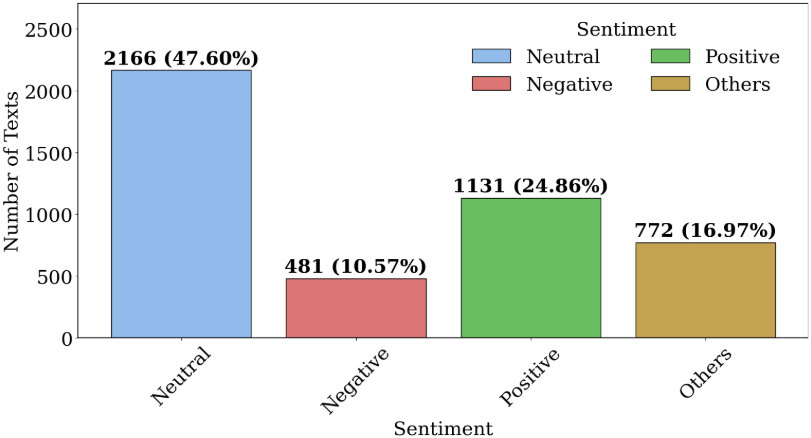}
  \caption{Number of texts in each sentiment class in the sample dataset before back translation.}
  \label{figSM1a}
\end{subfigure}
\hfill
\begin{subfigure}{\textwidth}
  \centering
\includegraphics[width=0.7\textwidth]{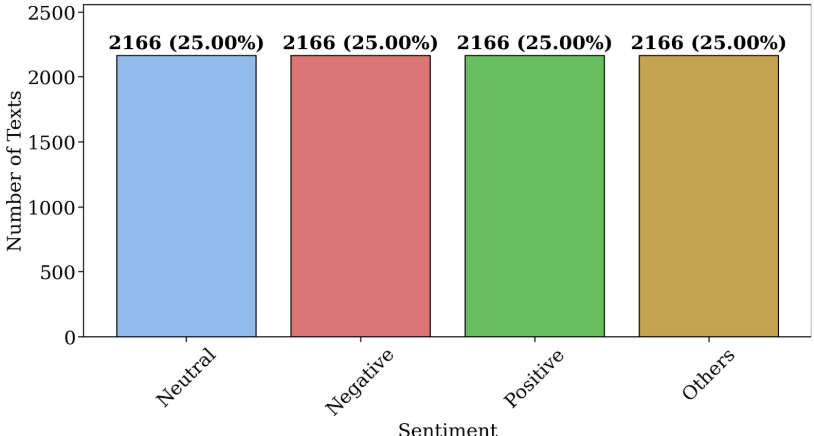}  
  \caption{Number of texts in each sentiment class in the sample dataset after back translation}
  \label{figSM1b}
\end{subfigure}
\caption{Number of texts in each sentiment class in the sample dataset before back translation and after back translation.} 
\label{figSM1}
\end{figure}

Looking at the distribution of the labels in Figure \ref{figSM1a}, there are 2,166 \emph{Neutral} texts, 1,131 \emph{Positive} texts, 772 \emph{Others} texts, and 481 \emph{Negative} texts. This uneven distribution could bias the models toward the more frequent classes. 
To address this issue, we applied back-translation as a data augmentation method \citep{sennrich2016improving}. Texts from the underrepresented classes were translated into German, Spanish, and French and then translated back into English. These languages were selected because large parallel corpora are available for training machine translation systems, which helps maintain semantic consistency during back-translation \citep{huggingface_languages}.
This back-translation process produced new versions that preserved the original sentiments, increased the number of samples in the less frequent classes, and helped the model learn all classes more evenly. After back-translation, the final annotated dataset included 8,664 texts, with a more balanced distribution among the four sentiment classes, as shown in Figure \ref{figSM1b}.



\subsection{Model Selection and Evaluation} \label{model_selection}

We reviewed the literature on NLP approaches for sentiment analysis in Section \ref{sec_Lit_rev_NLP_models}. Based on this review, we identified four main categories of methods: lexicon-based approaches, machine learning approaches, deep learning approaches with RNN architectures, and transformer-based architectures. From each category, we selected representative models to evaluate which approach performs best on our AAM dataset. We selected VADER from the lexicon-based category, Naive Bayes from the machine learning category, BiLSTM with attention and Text CNN combined with BiLSTM and attention from the RNN-based deep learning category, and RoBERTa, DeBERTa, and ModernBERT from the transformer-based deep learning category. This selection of models allowed us to evaluate which approach is most effective in detecting sentiment patterns in AAM-related discussions.


To evaluate which model should be used to annotate the main unannotated dataset, the annotated sample dataset was split into an 80\% training set (6,932 texts) and a 20\% testing set (1,732 texts). For the transformer-based models (RoBERTa, DeBERTa, and ModernBERT), fine-tuning involved taking models that had already been pre-trained on large general-language corpora and further training them on the 80\% training portion of our annotated AAM dataset. This process allowed the model parameters to adapt to the AAM-specific vocabulary, phrasing, and sentiment patterns. The neural architectures (BiLSTM with attention and Text CNN + BiLSTM with attention) were also trained on the same training data. The rule-based VADER model does not require fine-tuning because it relies on a predefined sentiment lexicon and rules. Similarly, the Naive Bayes classifier does not require fine-tuning and was simply trained on the training dataset. All models were evaluated on the 20\% testing set. We compared the test accuracies of the models and selected the one that achieved the highest performance on the test data. Table~\ref{tab:model-accuracy} reports the accuracies of the different models on the test dataset. The details of the different model types are described below.

\begin{table}[h]
\centering
\caption{Type of evaluated models and their accuracy on the test dataset.}
\label{tab:model-accuracy}
\begin{tabular}{lcc}
\hline
\textbf{Model} &  \textbf{Type} & \textbf{Test Accuracy}  \\
\hline
VADER  & Rule-based & 0.4461 \\ 
Naive Bayes & Classical Machine Learning & 0.7620 \\ 
BiLSTM + Attention & Deep Learning& 0.8376  \\
Text CNN + BiLSTM + Attention & Deep Learning& 0.8635 \\
RoBERTa  &  Deep Learning  & 0.9027\\
DeBERTa & Deep Learning& 0.9027  \\
ModernBERT & Deep Learning& 0.9391 \\
\hline
\end{tabular}
\end{table}

\subsubsection{VADER}

VADER (Valence Aware Dictionary and sEntiment Reasoner) is a rule-based model, in which sentiment is scored based on the existing sentiment lexicon \citep{Hutto_Gilbert_2014}. A sentiment lexicon consists of a annotated list of generally positive or negative sentiment words, such as "love," "nice," and "good" for positive sentiment and "hurt," "ugly," and "sad" for negative sentiment \citep{Hutto_Gilbert_2014}. VADER scores a text by taking its raw score and normalizing it to 1, resulting in a compound score that is used to label a text with a positive, negative, or neutral sentiment \citep{Hutto_Gilbert_2014}. The compound score is computed with the following equation. 

\begin{equation}
compound\_score = \frac{raw\_score}{\sqrt{{raw\_score}^2 + \alpha}}
\end{equation}

where $raw\_score$ represents the score calculated by VADER's lexicons. The normalization constant,  $\alpha$, is set to 15, and is used to approximate the maximum expected score \citep{Hutto_Gilbert_2014}. For compound scores less than -0.05 a negative sentiment is annotated, for compound scores between -0.05 and 0.05 a neutral sentiment is labeld, and for compound scores 0.05 a positive sentiment is annotated. Our annotated AAM sample dataset includes a fourth sentiment class, the \emph{Others} class; to run the original VADER model, we created a copy of our annotated dataset and deleted the data points with in the \emph{Others} class. The resulted in an accuracy of 0.5908. In order to run the model with our original annotated dataset including the  \emph{Others} class, we adjusted the scoring to include this new class. The adjusted scoring method is as follows: if the the compound score was less than -0.05, it was once again annotated \emph{Negative}; if the compound score fell between -0.05 and -0.02 or between 0.02 and 0.05, it was annotated \emph{Others}; if the compound score fell between -0.02 and 0.02, it was annotated \emph{Neutral}; and if the compound score fell above 0.05, it was once again annotated \emph{Positive}. This range was determined as the \emph{Others} class includes emotions such as confusion, curiosity, and surprise, which do not exhibit specifically positive or negative sentiments. The resulting accuracy for VADER when including the fourth sentiment class was 0.4461. Although VADER has performed with high accuracy when used to analyze text on social networks \citep{Hutto_Gilbert_2014}, its performance weakens when a fourth class of sentiments is added. VADER exhibited the lowest performance amongst the benchmark models as it cannot be "trained." As it is lexicon based, VADER scores text using its existing sentiment lexicon which cannot be changed or increased upon. 

\subsubsection{Naive Bayes}

The Naive Bayes classifier is a supervised machine learning model based on the Bayes' Theorem. The model undergoes supervised learning to assign sentiment classes to text \citep{amelia2026public}. We tested the effectiveness of the Naive Bayes Classifier in this AAM study as a benchmark model, as it is a more classical machine learning model. Naive Bayes was trained with our annotated AAM dataset to test its performance on social media text regarding AAM. The Naive Bayes classifier achieved an accuracy of 0.762 after training the model for the four sentiment classes. The transformer based Naive Bayes model performed higher than the rule-based VADER model, as the Naive Bayes model can be trained. However, as the Naive Bayes model is a more classical machine learning approach, it is expected to achieve lower accuracy compared to deep learning models. Therefore, we next implemented deep learning models to improve accuracy.

\subsubsection{BiLSTM with Attention}


The BiLSTM with attention model achieved an accuracy of 0.8376, outperforming the Naive Bayes classifier. Unlike Naive Bayes, which assumes independence among features, the BiLSTM architecture captures sequential dependencies and contextual relationships between words in a sentence \citep{wang2016attention}. The bidirectional structure processes the text in both forward and backward directions, allowing the model to incorporate information from both preceding and following tokens when learning contextual representations. The attention mechanism further enhances the model by assigning greater weight to words that contribute more strongly to the emotional meaning of the sentence \citep{zhou2016attention}. By focusing on these emotionally informative words, the model can better identify sentiment and emotion cues in the text, leading to improved classification performance compared to traditional machine learning approaches.



\subsubsection{Text CNN + BiLSTM with Attention}

To further improve performance, we evaluated a hybrid architecture combining a Text CNN for local n-gram feature extraction, a Bidirectional Long Short-Term Memory (BiLSTM) network for capturing contextual dependencies, and an attention mechanism for emphasizing sentiment-relevant words. The Text CNN first extracts local n-gram features from the input text using convolutional filters with kernel sizes of 3, 4, and 5, enabling the capture of trigram, four-gram, and five-gram lexical patterns. The feature maps produced by each kernel size are concatenated to form a unified local feature representation. This representation is then passed to the BiLSTM layer, which models bidirectional contextual dependencies across the sequence. An attention mechanism is applied on top of the BiLSTM outputs to assign higher weights to words that contribute more strongly to the emotional meaning of sentences \citep{kim2014cnn, yang2016hierarchical, zhou2016attention}. On the test dataset, the hybrid model achieved an accuracy of 0.8635, outperforming the standalone BiLSTM with attention model. This improvement indicates that incorporating CNN-based local feature extraction enhances the model’s ability to capture short-range emotional expressions and fine-grained lexical patterns, while the BiLSTM and attention components improve contextual understanding and feature weighting.

\subsubsection{RoBERTa}



RoBERTa achieved an accuracy of 0.9027. Its higher performance compared to RNN-based models is due to its ability to capture rich contextualized word embeddings and semantic dependencies \citep{liu2021probing}, which are important for sentiment classification. The model also identifies subtle contextual nuances and leverages knowledge gained from pretraining on large general-language corpora. 
These capabilities allow RoBERTa to better adapt to domain-specific sentiment patterns than RNN-based models.

\subsubsection{DeBERTa}

DeBERTa achieved 0.9027 accuracy, similar to RoBERTa. Although DeBERTa has advanced features, such as disentangled attention, that can better capture sentence structure \citep{he2020deberta}, in our dataset, overall sentiment patterns are already effectively captured by RoBERTa’s contextual embeddings. The additional improvements in DeBERTa do not provide much extra benefit for this task, so its performance ends up being very similar to RoBERTa. 

\subsubsection{ModernBERT}

ModernBERT is a recent transformer architecture that builds on BERT and RoBERTa with additional pretraining optimizations, including improved tokenization strategies, adaptive normalization techniques, and advanced regularization methods designed to improve generalization \citep{warner2025smarter}.
We apply ModernBERT by fine-tuning the pretrained encoder with an added linear classification head for the four sentiment classes. The model is trained end-to-end on the annotated dataset and achieves an accuracy of 0.9391. This was the highest performance among all models tested on our annotated dataset.

Since ModernBERT performed the best among all the models, the fine-tuned ModernBERT was used to label the sentiment of 301,459 unannotated texts. Each text was assigned a single label from the four-class sentiment taxonomy, \emph{Positive}, \emph{Negative}, \emph{Neutral}, or \emph{Others}, resulting in a fully annotated dataset ready for further analysis and modeling.

\subsection{Topic Modeling}

To identify the main topics expressed in the texts, we applied topic modeling to the annotated dataset. A topic refers to a semantically coherent cluster of words and phrases that collectively represent a recurring topic or subject discussed in the text data. The annotated dataset was divided into four subsets based on predicted sentiment classes. Each subset was analyzed separately to uncover the dominant topics within each emotion. For each sentiment subset, we initially experimented with alternative unsupervised approaches, including n-gram based representations (bigrams and trigrams) and word cloud visualizations, to explore dominant lexical patterns. However, these approaches primarily produced isolated or highly frequent word clusters that lacked clear semantic structure and did not adequately capture contextual relationships within the text. Therefore, we applied an LDA \cite{batool2024enhanced} for topic modeling. LDA provides a probabilistic framework that identifies latent topics within the text by grouping words into coherent topics based on their co-occurrence patterns across documents. As a result, it is more effective than frequency-based methods for discovering meaningful and interpretable topics.

The complete workflow of LDA topic modeling is summarized in Algorithm~\ref{alg:lda_pipeline}. Before topic modeling, each text was preprocessed using the spaCy NLP library \cite{spacy}. Texts were converted to lowercase, tokenized, and lemmatized to reduce words to their canonical forms. Punctuation, non-alphabetic tokens, standard English stopwords, and a set of domain-independent conversational terms (e.g., ``lol'', ``okay'', ``nice'') that do not contribute to semantic topic interpretation were removed. Tokens shorter than three characters were also excluded. The cleaned tokens were then rejoined to form a bag-of-words representation suitable for LDA. The modeling process began with the construction of a text--term matrix using count-based vectorization, where each document is represented by the frequency of its constituent terms. Terms appearing in more than 85\% of documents were removed because such extremely frequent words are typically non-informative for distinguishing between topics. These words tend to occur uniformly across the corpus and therefore do not contribute meaningful variation, often leading to blurred or overlapping topic representations. Additionally, terms appearing in fewer than 20 documents were excluded to reduce sparsity in the document--term matrix and to eliminate rare or idiosyncratic words. Such infrequent terms often arise from noise or highly specific expressions and do not provide reliable statistical evidence for learning consistent topic structures. Removing them helps stabilize the estimation of topic distributions by ensuring that topics are learned from sufficiently supported vocabulary. In addition, the vocabulary was capped at the 6,000 most frequent features to control the dimensionality of the input space. This reduction improves computational efficiency and helps prevent overfitting in the LDA model by focusing on the most informative and commonly occurring terms. LDA was then applied to this matrix to infer latent topic structures. The optimal number of topics was selected by training multiple candidate models and choosing the configuration that maximized the topic coherence score ($c_v$). The coherence score measures the degree of semantic consistency among the top words in a topic by evaluating their co-occurrence patterns in the corpus; higher coherence scores indicate more interpretable and semantically meaningful topics.

\begin{algorithm}[tb!]
\caption{LDA topic modeling}
\label{alg:lda_pipeline}
\begin{algorithmic}[1]
\REQUIRE Emotion-annotated texts $\mathcal{D} = \{d_1, d_2, \dots, d_N\}$
\ENSURE Topics, text--topic probabilities, representative texts, and insights

\STATE Filter texts belonging to a single sentiment class 
\STATE Preprocess texts using spaCy:
\STATE \hspace{0.5cm} lowercase, tokenize, lemmatize
\STATE \hspace{0.5cm} remove stopwords, short tokens, and domain-independent words
\STATE Construct a text--term matrix using count-based vectorization
\STATE Apply frequency filtering ($\text{min\_df}$, $\text{max\_df}$) and cap vocabulary size

\FOR{$k = 2$ to $10$}
    \STATE Fit LDA model with $k$ topics
    \STATE Extract top words per topic
    \STATE Compute topic coherence using the $c_v$ metric
\ENDFOR

\STATE Select optimal number of topics $k^*$ with highest coherence
\STATE Fit final LDA model using $k^*$ topics

\STATE Compute text--topic probability matrix $\Theta \in \mathbb{R}^{N \times k^*}$
\FOR{each text $d_i$}
    \STATE Assign dominant topic $\arg\max_j \Theta_{ij}$
\ENDFOR

\FOR{each topic $t_j$}
    \STATE Rank texts by $\Theta_{ij}$ in descending order
    \STATE Select top-$n$ texts as representative texts
\ENDFOR

\RETURN Topics, probabilities, and representative texts
\end{algorithmic}
\end{algorithm}


After training the final LDA model, we computed a text--topic probability matrix, where each row corresponds to a text and each column corresponds to a topic. Each entry represents the probability that a given text is associated with a specific topic. This probabilistic representation allows individual texts to express multiple topics with varying strengths, rather than enforcing a hard topic assignment. Table~\ref{tab:doc_topic_matrix} presents an illustrative subset of the text--topic probability matrix. Although only a small number of texts are shown for readability, the full matrix contains topic probability distributions for all texts in the corpus. Each text was assigned a dominant topic by selecting the topic with the highest probability score in its corresponding row of the text--topic probability matrix. This assignment was used for topic labeling and subsequent qualitative interpretation.

\begin{table}[h]
\centering
\caption{Text--topic probability matrix}
\label{tab:doc_topic_matrix}
\begin{tabular}{c c c c}
\hline
\textbf{Text ID} & \textbf{Topic 1} & \textbf{Topic 2} & \textbf{Topic 3} \\
\hline
1 & 0.0524 & 0.7639 & 0.1837 \\
2 & 0.0132 & 0.9023 & 0.0846 \\
3 & 0.0082 & 0.3819 & 0.6099 \\
4 & 0.4998 & 0.4757 & 0.0245 \\
5 & 0.0067 & 0.9869 & 0.0064 \\
\ldots & \ldots & \ldots & \ldots \\
\hline
\end{tabular}
\end{table}


To interpret the learned topics, representative texts were selected by ranking documents according to their topic probability scores and extracting the highest-scoring examples, as outlined in Algorithm~\ref{alg:lda_pipeline}. For each topic, the texts were ranked according to their probability of belonging to that topic and the top-ranked texts were selected as representative examples. These representative texts are necessary because they provide contextual evidence of each topic and helping validate that the learned structures correspond to meaningful semantic topics rather than only statistical word groupings. Finally, the LDA pipeline produced a set of dominant topics characterized by their most representative keywords and high-probability texts. 





\section{Results}

This section presents the results of our analysis, including performance evaluation of ModernBERT, sentiment prediction results by the model, topic modeling outcomes across different sentiment classes, and a cross-sentiment comparative analysis. Finally, we summarize key findings and discuss actionable strategies for converting \emph{Negative}, \emph{Neutral}, and \emph{Others} sentiments into \emph{Positive} sentiment.


\subsection{Performance of Sentiment Classification Model}

\begin{figure}[tb!h] 
    \centering
    \includegraphics[width=12cm,height=10cm]{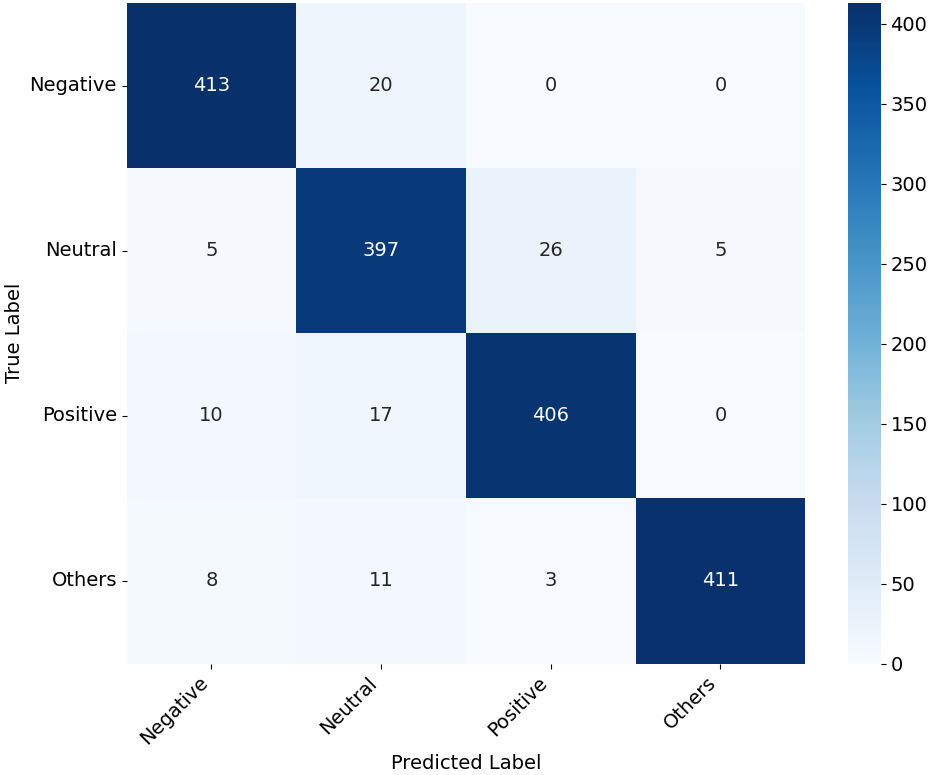}
    \caption{Confusion matrix of ModernBERT predictions on the test dataset} \label{confusion_matrix}
\end{figure}

As mentioned in Section~\ref{model_selection}, ModernBERT achieved the highest accuracy (0.9391) in the test data set and was selected to annotate the entire dataset. The detailed performance metrics are presented in Table~\ref{tab:class_metrics}. The model achieves high scores across all evaluation metrics, with accuracy, precision, recall, and F1-score all around 0.94. These high values indicate that the model performs reliably and makes accurate predictions across different sentiment classes. The corresponding confusion matrix is shown in Figure \ref{confusion_matrix}. The matrix shows that most predictions fall on the diagonal, meaning the model correctly assigns each text to its true sentiment class. The smaller off-diagonal values represent cases where the model assigned the wrong class. The low number of such errors indicates that the model distinguishes well between the different sentiment classes.

\begin{table}[h]
\centering
\caption{Performance metrics of ModernBERT on the test set.}
\label{tab:class_metrics}
\begin{tabular}{lccc}
\toprule
\textbf{Class} & \textbf{Precision} & \textbf{Recall} & \textbf{F1-score} \\
\midrule
Negative & 0.9472 & 0.9538 & 0.9505 \\
Neutral  & 0.8921 & 0.9169 & 0.9043 \\
Positive & 0.9333 & 0.9370 & 0.9351 \\
Others   & 0.9882 & 0.9492 & 0.9683 \\
\midrule
\textbf{Macro Average} & \textbf{0.9402} & \textbf{0.9392} & \textbf{0.9396} \\
\bottomrule
\end{tabular}
\end{table}

\subsection{Sentiment Prediction Results}

\begin{figure}[tb!h] 
    \centering
    \includegraphics[width=12cm,height=7cm]{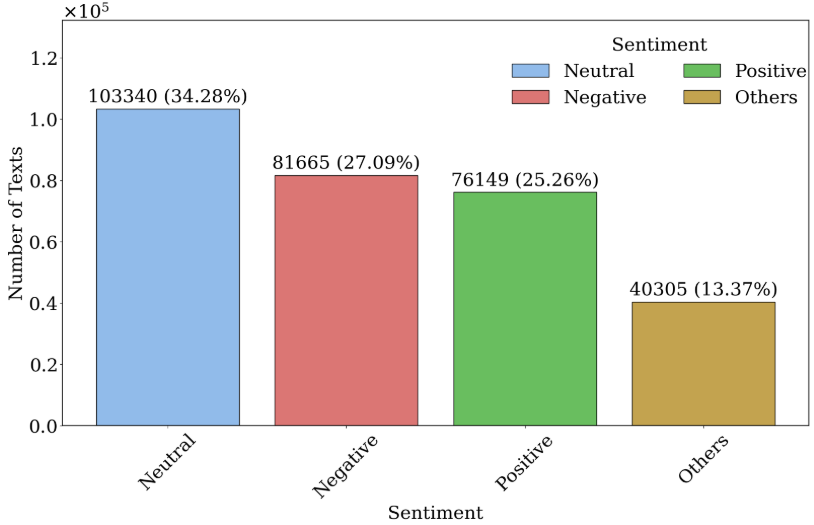}
    \caption{Sentiment distribution of the full AAM dataset (301,459 unannotated texts) predicted by ModernBERT.} \label{full_predicted}
\end{figure}

\begin{figure}[tb!h] 
    \centering
    \includegraphics[width=12cm,height=7cm]{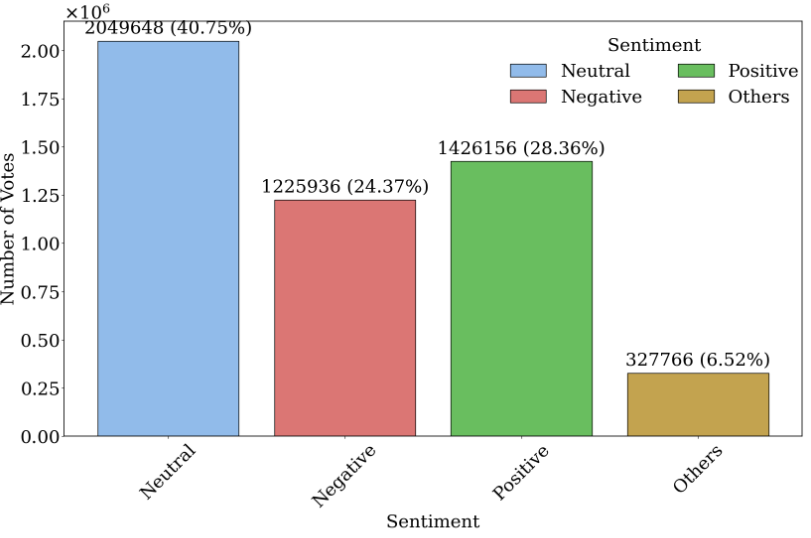}
    \caption{Distribution of net vote scores (upvotes minus downvotes) across predicted sentiment classes.} \label{votes}
\end{figure}

ModernBERT was then applied to the dataset of 301,459 unannotated texts to obtain sentiment labels. The resulting sentiment distribution is shown in Figure~\ref{full_predicted}. The distribution indicates that the Neutral class accounts for 34.28\%, followed by Negative (27.09\%), Positive (25.26\%), and Others (13.37\%). In addition to the textual data, post and comment-level engagement information was also collected in the form of upvotes and downvotes, defined as the net vote score (upvotes minus downvotes). Based on the predicted sentiment labels, the total vote counts associated with each sentiment category were analyzed. Figure~\ref{votes} presents this distribution. Out of a total of 3,746,006 votes, 40.75\% correspond to Neutral texts, 28.36\% to Positive texts, 24.37\% to Negative texts, and 6.52\% to Others. When looking at the posted texts, most content is neutral, with slightly more negative sentiment than positive, showing a generally balanced but mildly critical opinion on AAM. However, in terms of votes, users tend to engage more with neutral and positive posts than with negative or other classes, suggesting that while concerns exist, overall interaction is more supportive and optimistic.

\subsection{Topic Modeling Results}

To identify the underlying topics driving public sentiment, topic modeling was performed separately within each sentiment class. First, ModernBERT was used to assign sentiment labels (\emph{Positive}, \emph{Negative}, \emph{Neutral}, and \emph{Others}) to each text. Based on these labels, the dataset was partitioned into four sentiment-specific subsets. The LDA was then applied independently to each subset to extract latent topics within each sentiment class. This process identified six topics in the \emph{Positive} class, six topics in the \emph{Negative} class, four topics in the \emph{Neutral} class, and four topics in the \emph{Others} class. Within each sentiment class, the LDA model assigns a probability to every topic, representing the proportion of texts associated with that topic in that class. These topic probabilities are normalized by the LDA model such that, within each sentiment class, they sum to 1, thereby representing the relative distribution of discussions across topics in that class.

\subsubsection{Topics Associated with Positive Sentiment}

\begin{figure}[tb!h] 
    \centering
    \includegraphics[width=16cm,height=8cm]{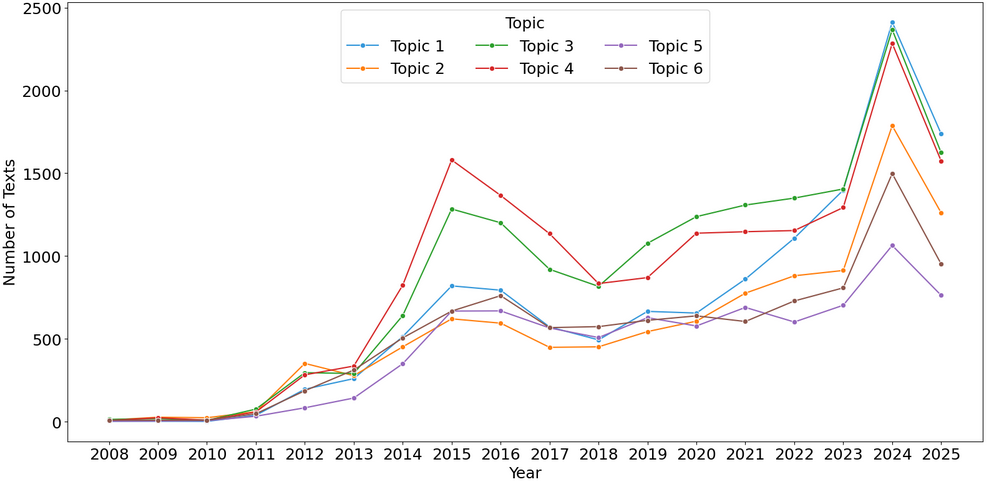}
    \caption{Temporal distribution of the six topics within the \emph{Positive} sentiment class from 2008 to mid-2025. 
    } \label{pos_topic_year}
\end{figure}

The \emph{Positive} sentiment class includes Topic 1 (Drone Operations and Applications), Topic 2 (Drone Software and Autonomous Systems), Topic 3 (Personal Experiences of Using Drones), Topic 4 (Drone Hardware and Flight Control), Topic 5 (Capability of Electric Aircraft), and Topic 6 (Drone Delivery and Future Urban Mobility). The temporal distribution of texts across the six topics within this class from 2008 to mid-2025 is presented in Figure~\ref{pos_topic_year}. Since the data for 2025 was collected only up to mid-year, a  dip is observed compared to 2024 due to incomplete yearly coverage. In Figure~\ref{pos_topic_year}, a noticeable surge across all topics is observed in 2015. This period corresponds to a phase of rapid development in the drone sector, during which the U.S. commercial drone industry began to emerge, military drone exports increased, technological advancements accelerated, and public safety concerns became widely debated \citep{michel2016drone}. 

A similar overall increase is also observed in 2024 across all topics. Topic 4 (Drone Hardware and Flight Control) and Topic 3 (Personal Experiences of Using Drones) dominate most of the timeline. This indicates that hands-on flying experiences and confidence in hardware reliability are the strongest drivers of positive sentiment. Topic 1 (Drone Operations and Applications) shows steady growth but begins to increase more sharply after 2020, reaching its highest level in 2024, indicating a shift toward greater attention to real-world drone applications. Topic 2 (Drone Software and Autonomous Systems) follows a more moderate growth pattern and does not dominate at any stage, suggesting that while autonomy is appreciated, it does not generate the same level of positive engagement as direct experience or hardware-related aspects. Topic 5 (Capability of Electric Aircraft) consistently records the lowest counts, indicating that futuristic electric aviation generates optimism but at a smaller scale compared to mainstream drone use. Topic 6 (Drone Delivery and Future Urban Mobility) shows gradual growth but remains below the dominant experiential and hardware-related topics. These topics are discussed in detail in the following paragraphs.

\paragraph{Topic 1: Drone Operations and Applications}

Discussions on this topic highlight positive perceptions of civilian and commercial drone operations (average probability: 0.1777). A subset of users on the platform talks about police drones that help with public safety, commercial drones used for real estate photography or property inspections, and professional UAVs that support search, rescue, or monitoring tasks. They emphasize the usefulness and efficiency of drones in these practical applications, noting how drones can make certain jobs easier and faster. This topic reflects that drones are seen as valuable tools that provide tangible benefits, helping some users feel more comfortable with and supportive of broader AAM deployment.

\paragraph{Topic 2: Drone Software and Autonomous Systems}


In this topic, a focus on intelligent and autonomous capabilities emerges (average probability: 0.1449), highlighting texts in which some users express interest in the software-driven and intelligent functionalities of drones. The discussions focus on autonomous navigation, flight planning algorithms, obstacle avoidance, and onboard intelligence that enables drones to operate with minimal human input. Users also express curiosity and admiration for the innovation behind these systems, noting how automation and smart decision-making enhance drone performance and reliability. The findings suggest that a subset of the platform participants view advanced software and autonomous features as key enablers of safe, efficient, and future-ready AAM operations.

\paragraph{Topic 3: Personal Experiences of Using Drones}


With an average topic probability of 0.2146, this topic captures how some users engage with drones through personal and recreational experiences. Conversations include first-person view (FPV) flights, recreational flying, beginner pilot experiences, and drone photography. Users share excitement, satisfaction, and pride in these experiences, highlighting how hands-on interaction makes drones enjoyable and approachable. In this topic, user-generated content on the platform suggests that personal enjoyment of drones can contribute to more positive attitudes toward aerial technology and can show greater openness to drones as part of everyday life and broader AAM operations.

\paragraph{Topic 4: Drone Hardware and Flight Control}

Representing the most prominent one, this topic (average probability: 0.2155) emphasizes hardware components and control mechanisms that allow drones to fly reliably. The texts in which some users discuss FPV rigs, flight controllers, telemetry, and signal stability, often reflecting on how these elements contribute to smooth, stable, and safe flight. Attention to these practical aspects in the texts suggests that some users appreciate drones as well-engineered and controllable machines. This topic highlights that solid hardware and flight control contribute to trust in drones.

\paragraph{Topic 5: Capability of Electric Aircraft}

This topic captures discussions among some users on the online platform regarding current and future-oriented developments in electric aircraft and eVTOL systems (average probability: 0.1148). The texts in this topic discuss electric propulsion, flight range, battery efficiency, and urban air mobility prototypes, often describing these AAM technologies and capabilities as sustainable and innovative alternatives to gas-driven automobile transportation. Overall, the discussions reflect enthusiasm for the potential of AAM to reduce emissions and urban congestion, and enable efficient mobility solutions. The generally positive sentiment among contributors suggests that some users view electric VTOL flight not only as technically feasible but also as a desirable and beneficial development for future cities.

\paragraph{Topic 6: Drone Delivery and Future Urban Mobility}

With an average topic probability of 0.1325, this topic reflects positive perceptions among some users about drone-enabled delivery and urban mobility innovations. The texts discuss last-mile logistics, autonomous delivery drones, and integration with ground transport and urban infrastructure, emphasizing efficiency, convenience, and futuristic appeal. The discussions express optimism that drones can improve package delivery, emergency supply transport, and broader urban logistics systems. The generally positive tone suggests that some users view AAM as a practical, innovative, and beneficial development with the potential to improve daily life and urban mobility.

\subsubsection{Topics Associated with Negative Sentiment}

\begin{figure}[tb!h] 
    \centering
    \includegraphics[width=16cm,height=8cm]{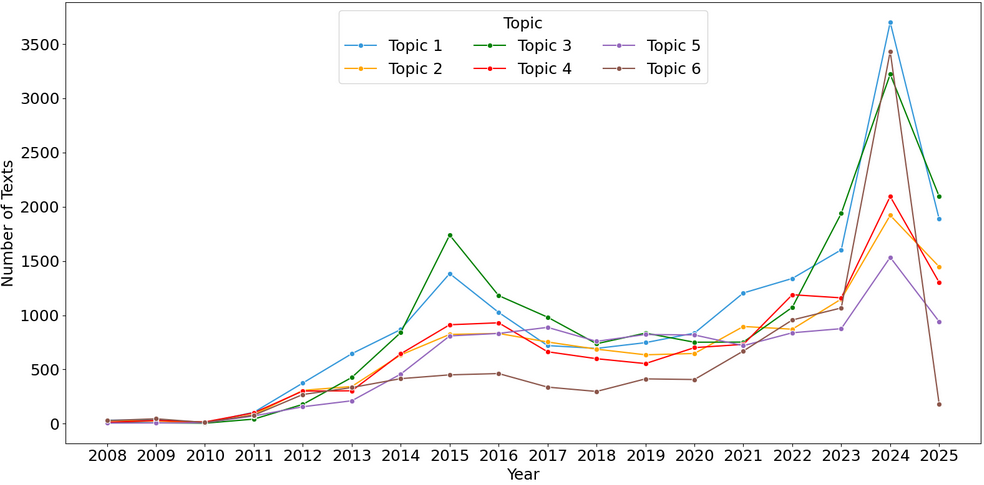}
    \caption{Temporal trends of negative drone-related discussions (2008–mid-2025).
    } \label{neg_topic_year}
\end{figure}

The Negative sentiment class includes Topic 1 (Safety Concerns about Drones), Topic 2 (Noise Concerns and Perceived Disturbance), Topic 3 (Regulation and Compliance), Topic 4 (Skill Gaps and Job Displacement), Topic 5 (Technical Constraints), and Topic 6 (Unethical Use of Drones). The trends in the volume of discussion of these six negative topics in social media, covering the period from 2008 to mid-2025, are presented in Figure~\ref{neg_topic_year}. Since the data for 2025 were collected only up to mid-year, a dip is observed compared to 2024 due to incomplete annual coverage.

All topics show a noticeable increase in 2015, except Topic 6 (Unethical Use of Drones), which remains relatively stable during that year. This spike in 2015 is largely driven by heightened regulatory activity and growing safety concerns. The U.S. Federal Aviation Administration (FAA) introduced mandatory drone registration rules in response to the rapid proliferation of unmanned aircraft systems and increasing reports of near-miss incidents between drones and manned aircraft in controlled airspace \citep{govtech2015faa}. These incidents included multiple close encounters involving commercial aircraft and drones, raising urgent concerns about airspace safety and enforcement gaps. Additional evidence from this period highlights drones as an emerging operational hazard in the absence of stricter regulatory frameworks, further amplifying public discourse on safety and compliance issues \citep{publicceo2016hazard}. A second major increase is observed after 2020, with the most pronounced rise occurring in Topic 6 (Unethical Use of Drones). This trend is primarily driven by the expanding role of drones in conflicts, such as the Russia–Ukraine war and the Israel–Gaza conflict. These developments have intensified public concern regarding the militarization of drone technology, surveillance misuse, and the ethical implications of autonomous and semi-autonomous systems.


Topic 3 (Regulation and Compliance) and Topic 1 (Safety Concerns) are the two most discussed topics on Reddit and Quora over the study timeline. The number of texts related to Topic 3 peaks strongly in 2015 and again rises sharply in 2023 and 2024, indicating that frustration with regulatory complexity becomes especially intense during periods of policy change or stricter enforcement. Topic 1 (Safety Concerns) follows a broadly similar trend to Topic 3 (Regulation and Compliance) but reaches the highest level in 2024 among all topics, suggesting that safety concerns become increasingly prominent as drone operations expand and become more visible in public airspace. Topic 6 (Unethical Use of Drones) remains moderate for many years but spikes sharply in 2024, nearly matching Topic 1 (Safety Concerns), showing that geopolitical tensions heavily influence recent negative sentiment. Topic 2 (Noise Concerns and Perceived Disturbance) remains relatively stable throughout the years and rarely leads, meaning it is a consistent but not dominant complaint. Topic 5 (Technical Constraints) maintains a mid-level presence throughout the timeline, reflecting steady skepticism about technological feasibility without extreme fluctuations. Topic 4 (Skill Gaps and Job Displacement) varies over time but does not dominate, indicating that workforce anxiety exists but is less central than concerns related to safety, regulation, and geopolitical issues. A detailed discussion of these topics is provided in the following paragraphs.

\paragraph{Topic 1: Safety Concerns about Drones}

With an average topic probability of 0.2090, this topic highlights concerns about the physical risks associated with drones and emerging concepts of flying vehicles. Participants in online discussions often express concerns about accidents, crashes, and the perceived lack of safety protections. It also includes privacy concerns related to cameras or drones operating near private residences. The insights indicate that safety remains a fundamental barrier to public acceptance as aerial technologies become more visible in everyday environments.

\paragraph{Topic 2: Noise Concerns and Perceived Disturbance}

This topic, with an average topic probability of 0.1484, highlights frustration among some users about the noise generated by drones and aerial vehicles. Many participants indicate that constant or high-frequency noise disrupts daily life and reduces quality of life in urban and suburban neighborhoods. The findings suggest that noise pollution is a significant concern that can negatively affect community acceptance if not adequately addressed.

\paragraph{Topic 3: Regulation and Compliance}

Regulatory-focused discussions emerge as a distinct topic (average topic probability: 0.2105), reflecting negative sentiment toward frameworks governing drone operations. The topic is characterized by frequent references to FAA regulations, Part 107 certification requirements, local drone ordinances, airspace classifications, flight restrictions, and distinctions between recreational and commercial activities. Rather than addressing technological capabilities, the discussions focus primarily on compliance-related challenges. Contributors commonly describe the regulatory environment as complex, restrictive, and administratively demanding for new entrants and small-scale operators.

In these texts, concerns are also raised regarding the perceived rigidity of airspace limitations, especially in urban areas where flight opportunities are constrained due to proximity to airports or controlled zones. Recurring issues include administrative procedures such as licensing exams, registration requirements, remote identification mandates, and difficulties in obtaining operational waivers. In some cases, frustration is expressed over perceived inconsistencies in enforcement or unclear distinctions between federal regulations and local ordinances.

The discussion reflects dissatisfaction directed not at drone technology itself but at the institutional frameworks governing its use. The findings suggest that regulatory complexity and compliance burdens contribute to negative perceptions of operational feasibility, with governance-related constraints acting as a structural barrier to adoption and limiting expansion opportunities for operators.

\paragraph{Topic 4: Skill Gaps and Job Displacement}

With an average topic probability of 0.1493, this topic captures discussions related to workforce readiness, skill acquisition, and employment uncertainty within AAM. The posts frequently reference technical competencies such as programming, engineering design, certification pathways, and flight training requirements, along with the qualifications needed to enter the industry and the competitiveness of the emerging job market. Beyond skill development, the topic also reflects concerns about the long-term stability of human roles within an increasingly automated ecosystem. Some contributors question whether advances in artificial intelligence, autonomous navigation systems, and AI-assisted flight operations may reduce demand for human pilots or operators, raising concerns about potential job displacement or role transformation as automation expands. The discourse combines cautious optimism about industry growth with uncertainty about employability and workforce sustainability. Although opportunities in the expanding AAM market are acknowledged, concerns about automation, required skill upgrades, and evolving job structures indicate that workforce adaptation remains a central issue in discussions surrounding AAM.

\paragraph{Topic 5: Technical Constraints} 

Technological limitations and feasibility challenges in drone and electric aviation systems are reflected in user-generated discussions on the platform (average topic probability: 0.1381). The texts in this topic frequently reference concerns about battery energy density, limited flight endurance, payload constraints, firmware malfunctions, and the broader viability of electric propulsion technologies. Rather than expressing opposition to the concept of AAM, the authors express skepticism about whether current technological capabilities are sufficient to support long-duration or large-scale operations. A recurring concern centers on limitations in battery performance, particularly the gap between the current energy storage capacity and the demands of sustained electric flight. Some discussions question the practicality of fully electric aircraft for commercial applications, citing weight-to-power trade-offs and insufficient range. Additional critiques focus on system reliability, including firmware issues, electronic speed controller failures, and software stability problems. Solar-powered solutions are also described as inefficient or unrealistic under current technological conditions. The discussion reflects technical doubt and feasibility concerns rather than ideological opposition. The sentiment stems from perceived engineering barriers and performance constraints that can limit scalability and operational viability.

\paragraph{Topic 6: Unethical Use of Drones} 

A topic centered on security concerns and geopolitical tensions associated with drone technology is reflected in user-generated discussions on the platform (average topic probability: 0.1448). The texts address the use of drones in military contexts, including swarm operations, autonomous weapon systems, and their application in battlefield scenarios. Drone swarms are often described as highly destabilizing or “scary” weapon technologies, with occasional comparisons drawn to nuclear deterrence in terms of their strategic impact. Beyond battlefield usage, concerns also extend to global power dynamics, including technological competition and manufacturing dominance, especially with respect to China's position in the drone market. These discussions frequently frame drone advancement not only as technological progress but also as a factor influencing military capability escalation and shifting geopolitical balances. The tone is characterized by anxiety, threat perception, and apprehension about potential future conflict intensification. The discourse highlights fears surrounding the militarization of drone technology rather than its civilian applications, with negative sentiment emerging from concerns about security risks, global instability, and the potential for rapid escalation enabled by autonomous or swarm-based systems. Such geopolitical anxieties can shape the interpretations of drone innovation by associating it more strongly with military threat than societal benefit.


\subsubsection{Topics Associated with Neutral Sentiment}

\begin{figure}[tb!h] 
    \centering
    \includegraphics[width=16cm,height=8cm]{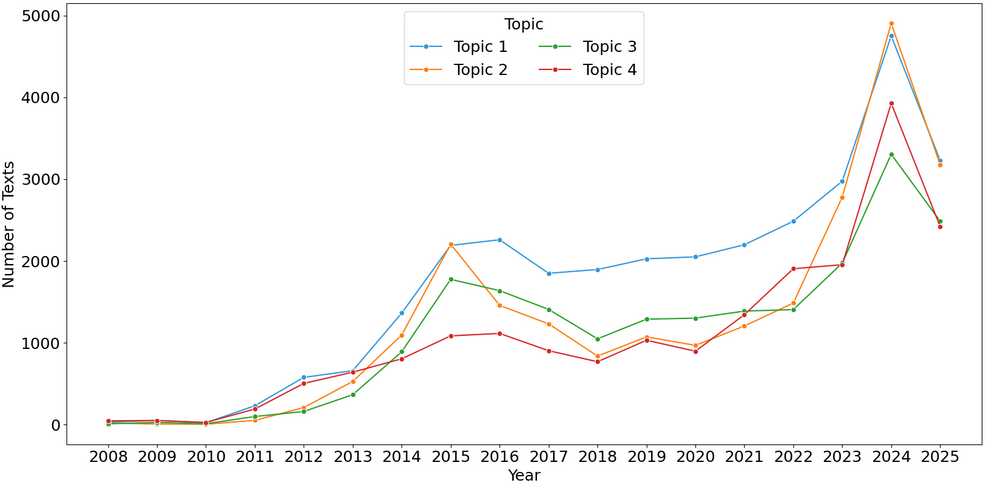}
    \caption{Temporal distribution of the four topics within the \emph{Neutral} sentiment class from 2008 to mid-2025. 
    }\label{neu_topic_year}
\end{figure}

The \emph{Neutral} class includes Topic 1 (Drone Industry, Workforce, and Production), Topic 2 (Drone Regulations and Airspace Management), Topic 3 (Drone Configuration, Stability, and Troubleshooting), and Topic 4 (Military and Defense Drones). Figure \ref{neu_topic_year} shows how these topics are discussed over the years. All topics exhibit noticeable increases between 2012 and 2016, followed by another rise after 2020. Topic 1 (Drone Industry, Workforce, and Production) remains the most consistently dominant topic throughout the study period. From 2011 onward, it generally records the highest counts, indicating that neutral discussions largely focus on industry operations, manufacturing, and workforce development. Topic 2 (Drone Regulations and Airspace Management) follows a different trend. Although it remains below Topic 1 for much of the timeline, it increases rapidly in 2023 and 2024, when it surpasses all other topics and reaches the highest peak in 2024. This suggests a recent shift in neutral discussions toward regulation and airspace management issues. Topic 3 (Drone Configuration, Stability, and Troubleshooting) exhibits a trend similar to Topic 1 (Drone Industry, Workforce, and Production) but generally remains at lower levels. In 2022, all other topics exceed Topic 3 (Drone Configuration, Stability, and Troubleshooting), indicating that hardware-focused discussions remain important but are not the dominant focus within neutral sentiment. Topic 4 (Military and Defense Drones) is relatively prominent during the early years, becomes more moderate over time, and rises again after 2020, suggesting renewed attention to defense-related drone discussions. These topics are discussed next.

\paragraph{Topic 1: Drone Industry, Workforce, and Production}

Industry-focused discussions about drone manufacturing and workforce structure are reflected with an average topic probability of 0.3145. The texts in this topic describe how UAVs and aircraft are produced, how production processes are organized, and the types of skills required in the workforce, including engineering knowledge, internships, and technical job roles. The content demonstrates an understanding of the operational and organizational aspects of the drone sector, focusing on human and technical resources without expressing explicit positive or negative sentiment.

\paragraph{Topic 2: Drone Regulations and Airspace Management}

Discussions about drone regulations and airspace rules appear with an average topic probability of 0.2421. Participants mention FAA requirements, licensing procedures, navigable airspace, property rights, and local ordinances, alongside explanations of how drones are incorporated into existing aviation frameworks. In some texts, law enforcement responsibilities and compliance considerations are also addressed. The general tone remains factual and informative, without evaluative judgment or emotional expression.

\paragraph{Topic 3: Drone Configuration, Stability, and Troubleshooting}

A topic centered on drone hardware and flight operations is reflected with an average topic probability of 0.2190. The texts address components such as motors, flight controllers, propellers, firmware, and battery systems, along with the configuration settings required for proper performance. The discussions also focus on stability, calibration, and troubleshooting, reflecting practical, experience-based knowledge of drone systems without expressing explicit positive or negative sentiment.

\paragraph{Topic 4: Military and Defense Drones}

With an average topic probability of 0.2244, this topic reflects neutral discussions on military and defense drone use. Users on the platform share news and information about surveillance drones, drones used in military operations, or international events involving drones. The content is mostly factual, focusing on capabilities and operations without including opinions or sentiments.

\subsubsection{Topics Associated with Others Sentiment}

\begin{figure}[tb!] 
    \centering
    \includegraphics[width=16cm,height=8cm]{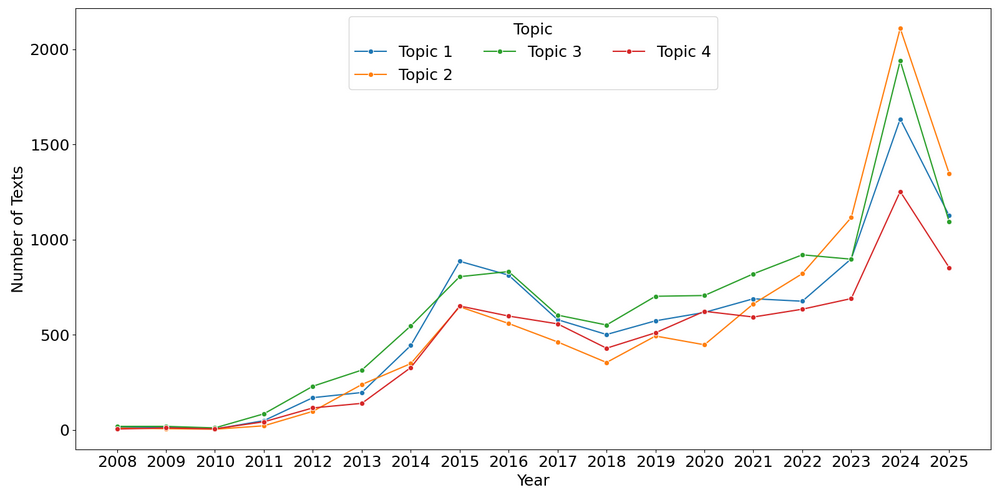}
    \caption{Temporal distribution of the four topics within the \emph{Others} sentiment class from 2008 to mid-2025. 
    } \label{others_topic_year}
\end{figure}

The \emph{Others} sentiment class includes Topic 1 (Emerging Drone Applications), Topic 2 (Drone Rules, Pilot Guidance, and Regulations), Topic 3 (Energy Systems of Drone), and Topic 4 (Engineering Educational Projects and Career Pathways). The trends of these topics are presented in Figure \ref{others_topic_year}. All topics exhibit noticeable increases in 2015, followed by another rise after 2020. Topic 3 (Energy Systems of Drone) is the dominant topic for most of the timeline, especially from 2010 to 2022, consistently recording higher counts than the other topics. This shows that technical and system-level discussions were historically the strongest area of engagement. However, from 2022 onward, Topic 2 (Drone Rules, Pilot Guidance, and Regulations) increases rapidly and overtakes all other topics by 2023, reaching the highest peak in 2024. This indicates a shift in recent years toward regulatory-focused discussions. Topic 1 (Emerging Drone Applications) grows steadily over time but remains below Topics 2 and 3, reflecting sustained but moderate interest. However, it was the most discussed topic among the four in the \emph{Others} sentiment class on Reddit and Quora in 2015. Topic 4 (Engineering Educational Projects and Career Pathways) remains the least discussed topic in nearly all years, suggesting that project- and career-related conversations are comparatively less prominent. The most notable change is the strong recent surge in Topic 2 relative to the more stable growth patterns of the other topics. These topics are discussed in detail in the following paragraphs.

\paragraph{Topic 1: Emerging Drone Applications}

Discussions reflecting surprise and curiosity about novel drone applications appear with an average topic probability of 0.2533. The conversations highlight emerging use cases such as urban air mobility for personal or public transport, drone-assisted delivery in densely populated cities, and monitoring remote or hard-to-reach areas. On platforms like Reddit and Quora, users often respond to these applications by reflecting on their unexpected nature and considering how such innovations could shape daily life, transform industries, and influence the delivery of public services. Alongside this interest, some discussions also raise accessibility considerations regarding the possibility that AAM services may primarily serve high-income neighborhoods, reflecting broader reflections on how the benefits of emerging air mobility systems may be distributed across society.

\paragraph{Topic 2: Drone Rules, Pilot Guidance, and Regulations}

A topic reflecting curiosity and surprise about drone laws and operational rules appears with an average topic probability of 0.2512. Discussions often center on local and national regulations, FAA guidance, licensing requirements, and the legality of drone flights in parks, urban areas, and event spaces. Many posts take the form of questions about proper flight practices, compliance obligations, and expectations for new pilots, reflecting an effort to understand the legal framework and ensure safe operation. The discourse remains factual and exploratory, highlighting how users interpret and learn about the rules that govern the use of drones.

\paragraph{Topic 3: Energy Systems of Drone}

With an average topic probability of 0.2889, this topic shows curiosity and analytical thinking about drones in the context of vehicles, aircraft, and power systems. People discuss drone motors, battery and energy efficiency, propulsion systems, and comparisons with other vehicles like cars or planes. Some texts explore technical questions about performance, efficiency, or hybrid systems. The tone is neutral and exploratory, highlighting public engagement with how drones work in broader transportation and energy contexts.

\paragraph{Topic 4: Engineering Educational Projects and Career Pathways}

This topic, with an average topic probability of 0.2065, reflects curiosity and confusion about engineering education, projects, and career planning. Discussions include student projects, university programs in robotics, mechatronics, or drone-related engineering, and internships or job experiences. The texts posted by the users share challenges in choosing courses, managing projects, or deciding on career paths. The discussions are informative and reflective, showing how people explore their options, learn from experience, and navigate decisions related to drones, engineering, and technology careers.

\subsubsection{Dominant Topics Within and Across Sentiment Classes}

\begin{figure}[tb!] 
    \centering
    \includegraphics[width=16cm,height=7cm]{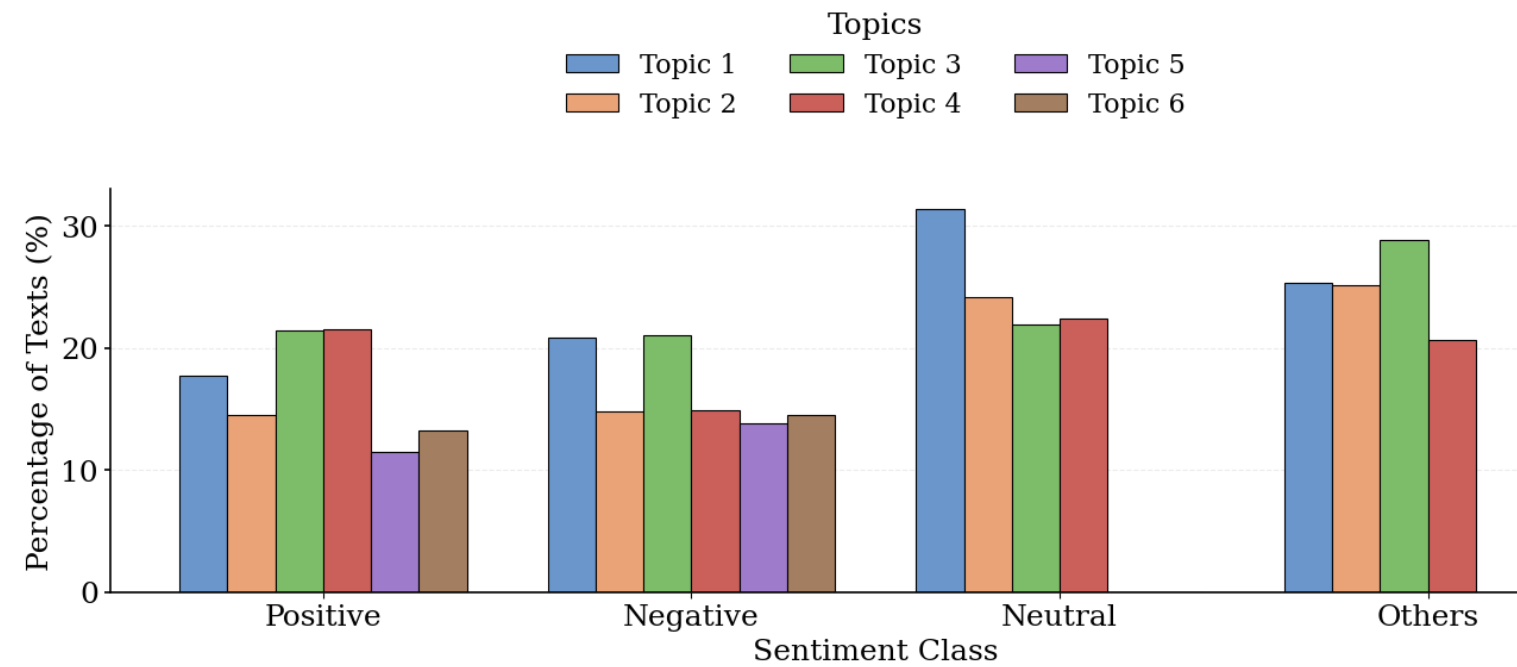}
    \caption{Topic distributions across the four sentiment classes, showing the percentage of texts from the 301,459 texts associated with each topic within each class. 
    } \label{topics_emotions_count}
\end{figure}

This section provides a consolidated overview of the dominant topics identified across all sentiment classes. The purpose is to summarize the most prominent topics emerging from the topic modeling results before proceeding to cross-sentiment comparison and interpretation. The corresponding distributions are visually illustrated in Figure~\ref{topics_emotions_count}. All percentages represent the proportion of the total dataset (301,459 texts) assigned to each topic within each sentiment class. 

Across the \emph{Positive} class, discourse is primarily centered on \emph{Drone Hardware and Flight Control} (21.55\%) and \emph{Personal Experiences of Using Drones} (21.46\%), with both topics exhibiting nearly equal prominence. This suggests that positive sentiment is largely associated with direct interaction, usability, and experiential engagement with drone technologies. In the \emph{Negative} class, \emph{Regulation and Compliance} (21.05\%) slightly exceeds \emph{Safety Concerns about Drones} (20.90\%), indicating that negative discourse is jointly shaped by regulatory constraints and perceived safety risks rather than a single dominant concern. The \emph{Neutral} class is strongly dominated by \emph{Drone Industry, Workforce, and Production} (31.45\%), reflecting a predominantly descriptive and informational discourse focused on industrial structure, workforce dynamics, and production-related aspects rather than evaluative opinions. In contrast, the \emph{Others} class is led by \emph{Emerging Drone Applications} (28.89\%), highlighting exploratory and forward-looking discussions that emphasize novel use cases and future possibilities.

Aggregating across all 20 topics across the four sentiment classes, the most dominant topics are \emph{Drone Industry, Workforce, and Production} (31.45\%), \emph{Emerging Drone Applications} (28.89\%), \emph{Drone Hardware and Flight Control} (21.55\%), \emph{Personal Experiences of Using Drones} (21.46\%), and \emph{Regulation and Compliance} (21.05\%). Overall, these results suggest that public discourse on drones is shaped by a combination of industrial and structural considerations, forward-looking and emerging application topics, regulatory and safety-related concerns, and hands-on technological engagement.

\subsection{Cross-Sentiment Topic Analysis}

Following the topic-level analysis, the next objective is to identify the main concerns within the \emph{Negative}, \emph{Neutral}, and \emph{Others} sentiment classes that can be translated into strategies for improving public perceptions of AAM. To support this objective, it is necessary to determine whether the 20 topics represent separate discussions or different sentiment-based expressions of the same underlying issue.

\begin{figure}[tb!h]
\centering

\begin{minipage}{0.48\textwidth}
    \centering
    \includegraphics[width=\textwidth]{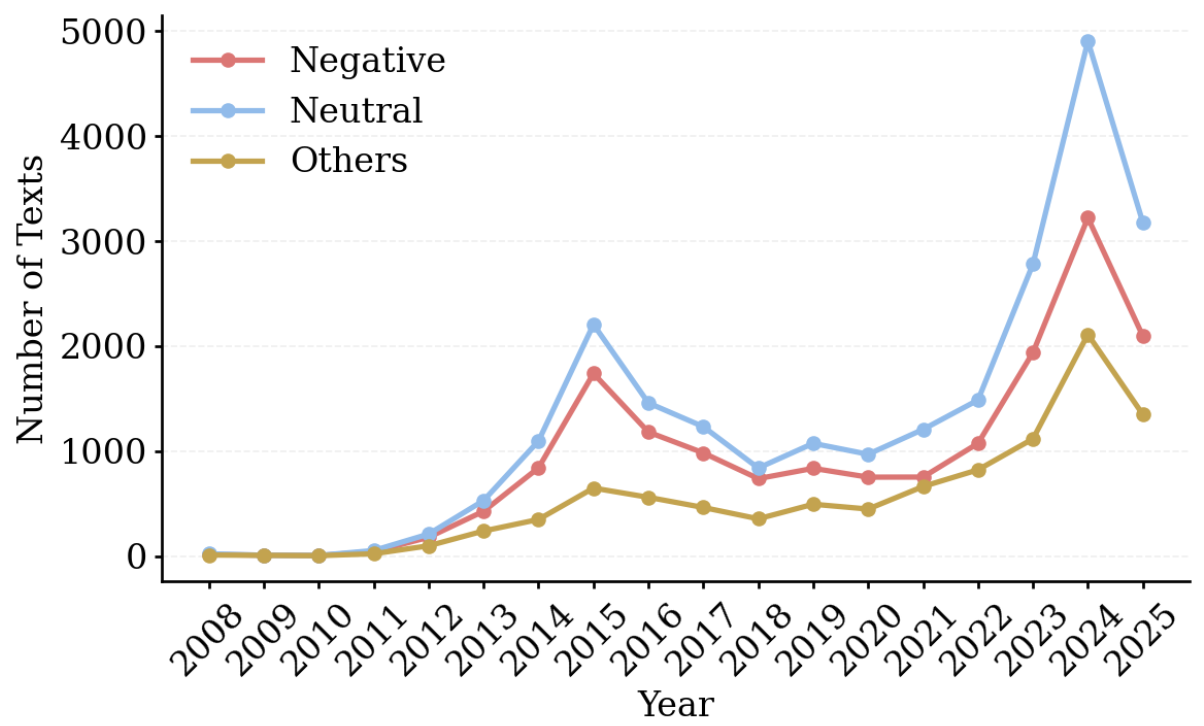}
    \caption{Yearly sentiment trends in the Regulation and Compliance cluster.}
    \label{fig:regulation_compliance}
\end{minipage}
\hfill
\begin{minipage}{0.48\textwidth}
    \centering
    \includegraphics[width=\textwidth]{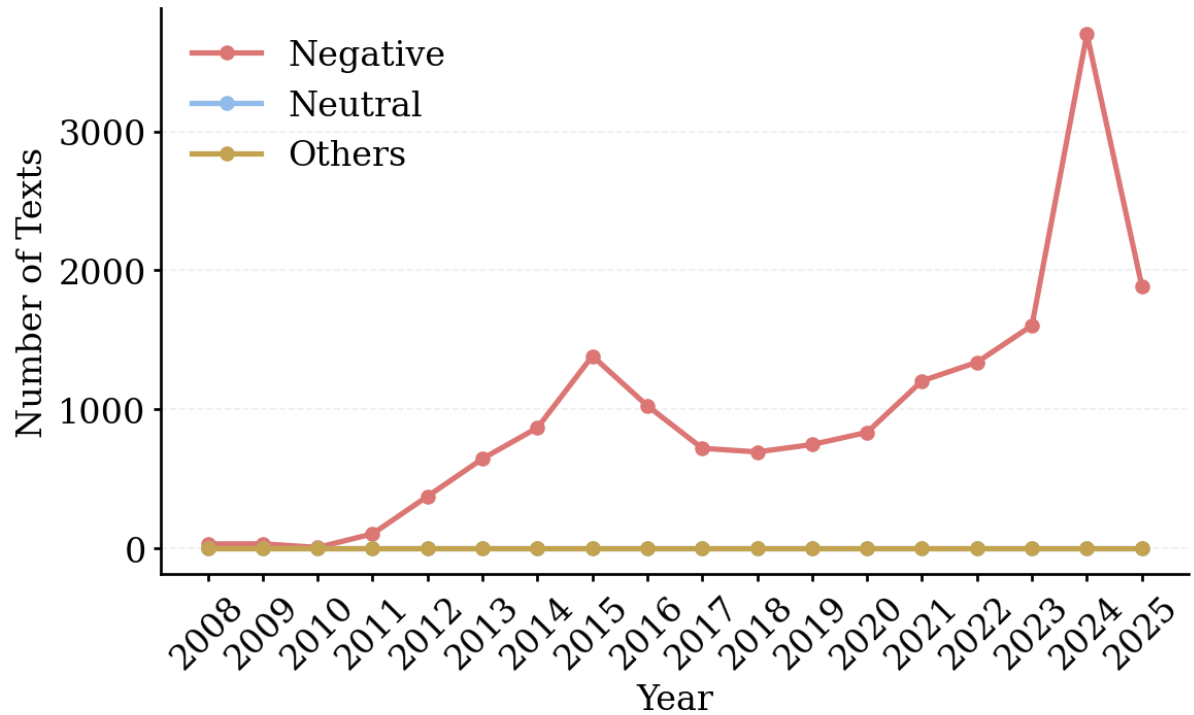}
    \caption{Yearly sentiment trends in the Safety and Operational Risks cluster.}
    \label{fig:safety}
\end{minipage}

\end{figure}

\begin{figure}[tb!]
\centering

\begin{minipage}{0.48\textwidth}
    \centering
    \includegraphics[width=\textwidth]{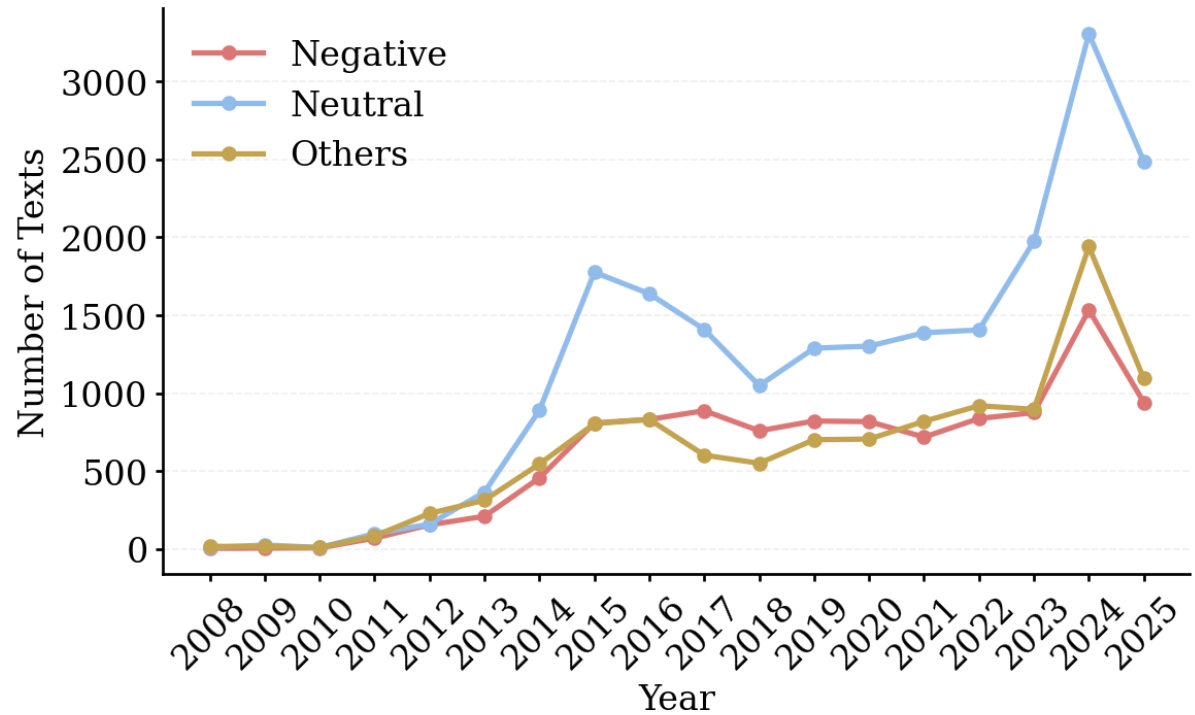}
    \caption{Yearly sentiment trends in the Technical Performance of Drones cluster.}
    \label{fig:technical_hardware_flight_performance}
\end{minipage}
\hfill
\begin{minipage}{0.48\textwidth}
    \centering
    \includegraphics[width=\textwidth]{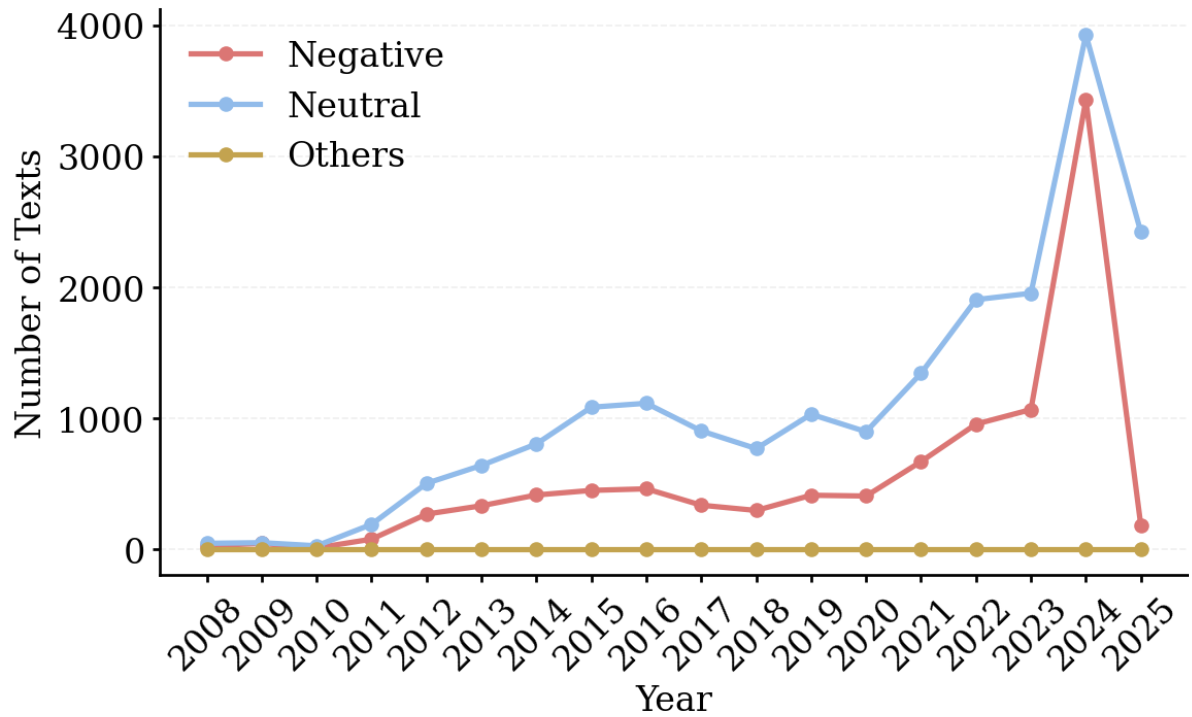}
    \caption{Yearly sentiment trends in the Military, Geopolitics, and Defense cluster.}
    \label{fig:military_geopolitics_defense}
\end{minipage}

\end{figure}

\begin{figure}[tb!]
\centering

\begin{minipage}{0.48\textwidth}
    \centering
    \includegraphics[width=\textwidth]{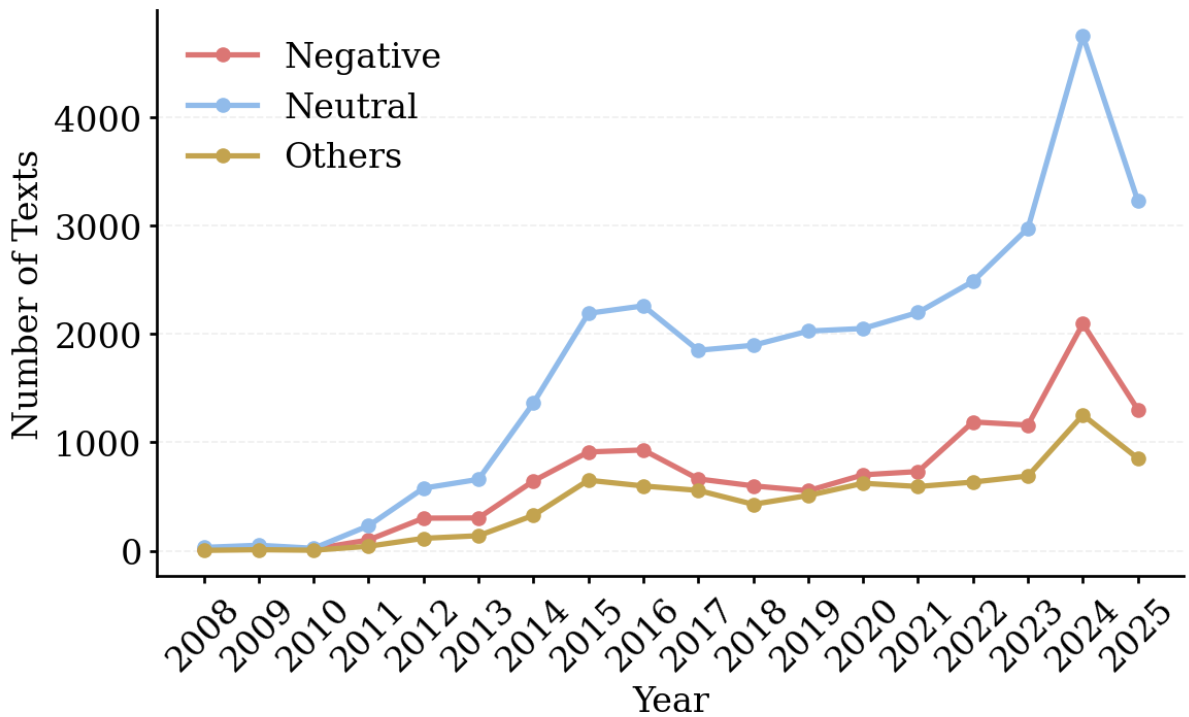}
    \caption{Yearly sentiment trends in the Workforce, Skill Development, and Career Pathways cluster.}
    \label{fig:workforce_career_skill}
\end{minipage}
\hfill
\begin{minipage}{0.48\textwidth}
    \centering
    \includegraphics[width=\textwidth]{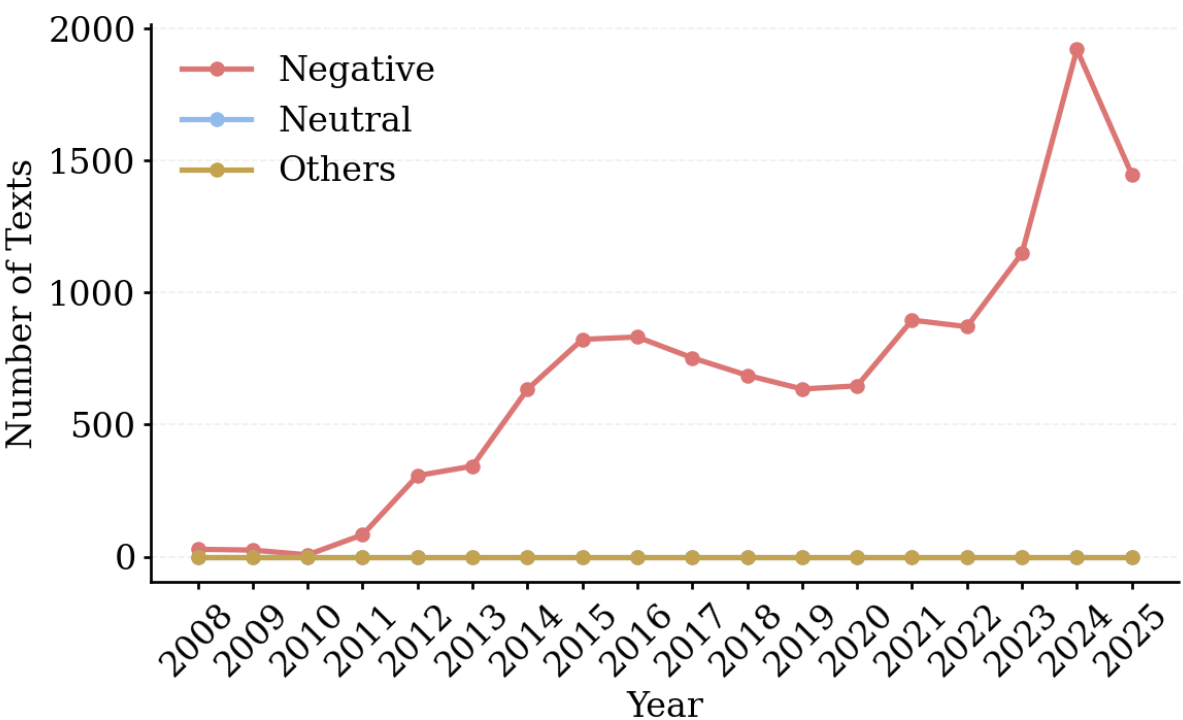}
    \caption{Yearly sentiment trends in the Noise and Disturbance cluster.}
    \label{fig:noise}
\end{minipage}

\end{figure}

The temporal patterns shown in Figures \ref{fig:regulation_compliance}--\ref{fig:noise} help clarify this relationship. In the Regulation and Compliance cluster, for example, the \emph{Negative}, \emph{Neutral}, and \emph{Others} trends follow similar increases and decreases over time, as shown in Figure \ref{fig:regulation_compliance}. This indicates that the central issue remains consistent across sentiment classes, while the way users discuss it changes. The same subject may be expressed as a concern in \emph{Negative} texts, as information-seeking in \emph{Neutral} texts, or as general discussion in the \emph{Others} class. Similar temporal relationships are also observed in the Safety and Operational Risks, Technical Performance of Drones, Military, Geopolitics, and Defense, Workforce, Skill Development, and Career Pathways, and Noise and Disturbance clusters, as shown in Figures \ref{fig:safety}--\ref{fig:noise}. These patterns suggest that sentiment differences often reflect changes in viewpoint and expression rather than entirely different discussion topics.

Based on this observation, related topics are grouped into broader clusters when they address the same underlying issue. This grouping is used only for higher-level interpretation and does not alter the original topic-level analysis. Accordingly, six focus clusters are identified in Table \ref{tab_six_focus}: (i) Regulation and Compliance, (ii) Safety and Operational Risks, (iii) Technical Performance of Drones, (iv) Military, Geopolitics, and Defense, (v) Workforce, Skill Development, and Career Pathways, and (vi) Noise and Disturbance.

\clearpage

\begin{table}[h] 
\centering
\caption{Six clusters of concern across \emph{Positive}, \emph{Negative}, \emph{Neutral}, and \emph{Others} sentiment classes.} \label{tab_six_focus}
\begin{tabular}{p{0.9cm}p{3.5cm} p{3cm} p{3cm} p{3cm}}
\hline
\textbf{Cluster No.} & \textbf{Cluster Name}    & \textbf{\emph{Negative} Class}                                   & \textbf{\emph{Neutral} Class}                                 & \textbf{\emph{Others} Class}                                  \\ \hline
\textbf{1} & \textbf{Regulation and Compliance}             & Topic 3 (Regulation and Compliance)                   & Topic 2 (Drone Regulations and Airspace Management)    & Topic 2 (Drone Rules, Pilot Guidance, Regulations)    \\ \hline
\textbf{2} & \textbf{Safety and Operational Risks}          & Topic 1 (Safety Concerns about Drones) &                                                        &                                                        \\ \hline
\textbf{3} & \textbf{Technical Performance of Drones} & Topic 5 (Technical Constraints)                          & Topic 3 (Drone Configuration, Stability, and Troubleshooting)  & Topic 3 (Energy Systems of Drone)            \\ \hline
\textbf{4} & \textbf{Military, Geopolitics, and Defense}     & Topic 6 (Unethical Use of Drones)                         & Topic 4 (Military and Defense Drones)                  &                                                        \\ \hline
\textbf{5} & \textbf{Workforce, Skill Development, and Career Pathways}           & Topic 4 (Skill Gaps and Job Displacement)                  & Topic 1 (Drone Industry, Workforce, Production)       & Topic 4 (Engineering Educational Projects and Career Pathways)     \\ \hline
\textbf{6} & \textbf{Noise and Disturbance}                 & Topic 2 (Noise Concerns and Perceived Disturbance)         &                                                        &                                                        \\ \hline
\end{tabular}

\end{table}

\begin{figure}[tb!] 
    \centering
    \includegraphics[width=16cm,height=8cm]{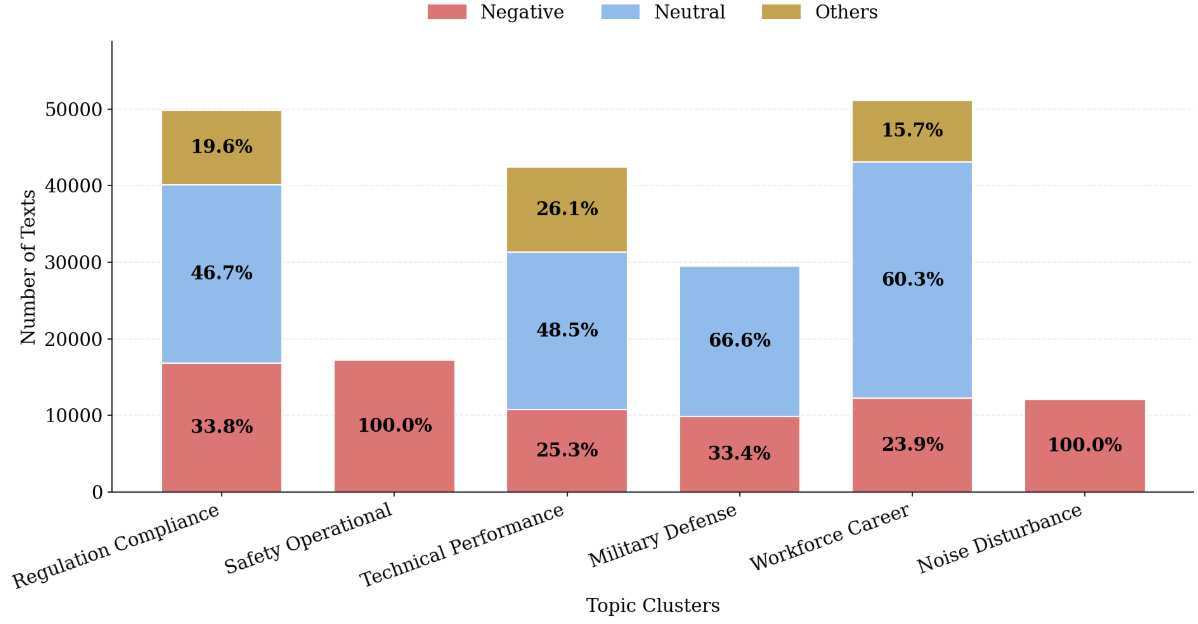}
    \caption{Distribution of AAM discussions across identified clusters of concern. 
    } 
    \label{cluster_total}
\end{figure}

Figure \ref{cluster_total} presents the total number of texts in each cluster, indicating how Reddit and Quora users engage in discussions on AAM. All reported cluster percentages are computed with respect to the entire dataset, comprising all texts across all clusters and sentiment classes. Workforce Career accounts for 25.29\% of the entire dataset, making it the most dominant cluster. This reflects the platform demographics, where users are generally younger and more actively engaged with emerging technologies. As a result, discussions related to jobs, skills, and career pathways are particularly prominent, as users seek to understand their positioning within the evolving AAM job market. Within this cluster, sentiment is mainly neutral (60.35\% of Workforce Career texts), followed by negative (23.94\%) and others (15.71\%), indicating a largely informational and exploratory discussion pattern. Next, Regulation Compliance cluster accounts for 24.64\% of the entire dataset, making it the second most prominent cluster. These discussions are often driven by users engaging with drone technologies and seeking clarification on FAA regulations, operational constraints, and certification requirements. Within this cluster, 46.68\% of Regulation Compliance texts are neutral, 33.76\% are negative, and 19.56\% belong to other sentiments, reflecting a mix of informational queries and concern-driven narratives. Technical Performance accounts for 20.99\% of the entire dataset, forming the third largest cluster. Within this cluster, sentiment is distributed as 48.53\% neutral, 25.34\% negative, and 26.14\% others, indicating a predominantly technical and discussion-oriented framing. Then, Military Defense cluster contributes 14.58\% of the entire dataset, ranking fourth among all clusters. Within this cluster, 66.57\% of Military Defense texts are neutral and 33.43\% are negative, with no other sentiment observed. Safety Operational accounts for 8.51\% of the entire dataset, and all associated texts are negative (100\% of Safety Operational texts), reflecting strong concern-driven discussions. Similarly, Noise Disturbance represents 5.98\% of the entire dataset, with 100\% of these texts being negative, indicating consistently critical sentiment in this cluster.

\subsection{Actionable Strategies for Addressing Key Concerns}

Based on the clusters of concern identified in Table \ref{tab_six_focus}, the objective is to prioritize areas whose mitigation would most effectively support a shift toward more positive public perception and broader adoption of drone technologies. Each cluster can be associated with actionable and evidence-based strategies that have already demonstrated effectiveness in related domains such as aviation, autonomous systems, and smart mobility.

\subsubsection{Strategies to Address Workforce, Skill Development, and Career Pathways Related Concerns in AAM}

Workforce, skill, and career-related concerns represent the most discussed cluster of concern, driven by uncertainty about required competencies, difficulty entering the field, fears of automation-induced job displacement, ambiguity in technical expectations such as programming, engineering design, certification pathways, flight training, and unclear career progression routes. To address these concerns and strengthen positive perceptions of AAM careers, the following four strategies can be implemented.

\vspace{0.2cm}

\noindent \textbf{i. Entry Clarity Through Structured and Transparent Career Pathways}: To begin with, many concerns originate at the entry stage due to unclear or fragmented information about available roles and required qualifications. Individuals often struggle to understand how to enter the AAM sector and what competencies are expected for different positions. To address this foundational issue, structured and transparent career pathways are essential. This can be achieved by clearly mapping role-specific pathways, for example distinguishing tracks for drone operators, software engineers, maintenance technicians, and system analysts, and organizing them into progressive levels such as beginner, intermediate, and advanced \citep{raza2025advanced,fein2012career}. Building on this structure, providing explicit guidance on qualifications, expected competencies, and career progression trends helps individuals make informed decisions about education and training pathways \citep{fein2012career,right2019future}. Furthermore, centralized digital platforms that integrate education, certification requirements, and employment pathways can act as navigation systems for learners. By connecting these elements into a single ecosystem, uncertainty at the entry stage is significantly reduced, thereby improving confidence in AAM career selection.

\vspace{0.2cm}

\noindent \textbf{ii. Job Readiness Through Experience-Based Training and Industry Integration}: Once entry pathways are clarified, the next challenge concerns preparedness for real-world employment. In particular, many individuals face a gap between theoretical knowledge and the practical competencies required in operational AAM environments. To bridge this gap, experience-based training must be embedded directly into learning systems. This includes simulation-based flight training, real-world project work, and supervised operational exercises, all of which allow learners to apply theoretical knowledge in realistic but controlled environments \citep{kelemen2023hybrid,van2020applying}. Through these mechanisms, individuals gradually build confidence and operational competence. In addition, job readiness is further strengthened when training is closely aligned with industry practice. Specifically, structured internships, entry-level placements, and project-based hiring pipelines ensure that learning outcomes match workforce expectations \citep{rohiat2025innovation}. As a result, the transition from education to employment becomes more direct, reducing concerns about competitiveness and employability.

\vspace{0.2cm}

\noindent \textbf{iii. Future Role Transformation Through Hybrid and Cross-Disciplinary Skill Development}: Beyond initial job readiness, concerns also shift toward the impact of automation and artificial intelligence on future employment. A key fear is that autonomous systems may replace human roles in AAM operations. However, this concern can be addressed by reframing the issue as a transition toward human--automation collaboration rather than replacement. Accordingly, training should emphasize roles that involve supervising autonomous operations, managing and interpreting data output, and making complex decisions that require human judgment \citep{chancey2020designing}. These responsibilities highlight that humans remain essential in safety-critical and oversight functions, even in highly automated systems. In parallel, promoting cross-disciplinary skill development further strengthens workforce adaptability. Encouraging individuals to combine flight operations knowledge with data analysis, software interaction, or system monitoring enables them to operate effectively in hybrid technological environments \citep{barari2026integrating}. This combination not only reduces fears of job displacement but also positions automation as a tool that enhances human capability and enables more specialized, higher-value roles.

\vspace{0.2cm}

\noindent \textbf{iv. Long-Term Career Sustainability Through Continuous Learning and Upskilling Systems}: Finally, even after successful entry and adaptation, long-term concerns remain regarding skill obsolescence in a rapidly evolving technological environment. As AAM technologies, regulations, and operational systems continue to develop, previously acquired skills may lose relevance over time. To address this, continuous learning and upskilling systems are necessary to support long-term workforce sustainability. This can be achieved by providing flexible learning opportunities such as short-term certification updates, refresher courses, and targeted training programs aligned with emerging technological and regulatory changes \citep{di2023future}. Moreover, ensuring that these opportunities are affordable and accessible across different career stages encourages sustained participation over time \citep{jenkins2021patterns}. In this way, continuous learning frameworks enable individuals to systematically update their skills in response to evolving demands, thereby reducing uncertainty and supporting long-term employability in the AAM sector.

\subsubsection{Strategies to Improve Regulation and Compliance in AAM}

Regulatory and compliance-related concerns represent a key barrier to the adoption of AAM and constitute the second most frequently discussed cluster of concern. Ongoing efforts by regulatory authorities, including the FAA and the European Union Aviation Safety Agency (EASA), aim to formalize operational frameworks, certification pathways, and airspace integration strategies for AAM systems \citep{ButterworthHayes2024EASA_AAM,EASA_UAM_2026,FAA2023UAMConOps}. However, despite these developments, public concerns persist in our social media dataset, primarily driven by perceptions of complex regulatory structures, administrative burdens, and limited clarity in approval processes. To address these concerns and support AAM integration, several targeted strategies can be implemented, as outlined below.

\vspace{0.2cm}

\noindent \textbf{i. Simplified and Role-Based Regulatory Frameworks}: A primary source of regulatory concern is the difficulty users face in interpreting complex aviation rules that are often written in technical or legal language. In addition, different types of AAM users have different responsibilities and levels of operation, meaning they do not all need the same regulatory information. For example, a beginner pilot, a commercial operator, and an infrastructure inspector each follow different rules and approval processes. For this reason, regulatory systems can be redesigned into simplified and role-based frameworks that present only the rules relevant to each user category, making the information easier to understand and apply \citep{toivonen2025transforming,sun2025examination}. Building on this, step-by-step compliance pathways can be introduced to guide users through processes such as registration, certification, and operational approval. This reduces complexity and helps users follow regulations in a clear sequence \citep{wallace2017evaluating}. In addition, interactive digital compliance tools can provide real-time guidance on required documents and procedures, further simplifying regulatory interpretation and reducing confusion.

\vspace{0.2cm}

\noindent \textbf{ii. Digitalized and Automated Compliance Enforcement Systems}: Beyond the simplification of rules, another critical strategy is the automation of compliance verification to reduce the need for manual interpretation of regulatory requirements. In this approach, regulatory constraints are embedded directly into operational systems through digital platforms that support automated airspace validation, real-time flight authorization, and geofencing alerts that prevent entry into restricted zones \citep{shrestha2021survey}. Rather than requiring operators to independently interpret airspace classifications or legal constraints, compliance is enforced dynamically within the operational environment itself. This integration of regulation into software-based control systems reduces human error, lowers administrative burden, and increases operational confidence. As a result, compliance becomes an inherent feature of AAM system design rather than an external procedural requirement.

\vspace{0.2cm}

\noindent \textbf{iii. Adaptive Airspace Management and Dynamic Operational Access}: In addition to regulatory simplification and automation, concerns regarding airspace rigidity can be addressed through adaptive airspace management frameworks. Traditional fixed airspace restrictions can limit operational flexibility, particularly in dense urban environments where demand for low-altitude mobility is high. To overcome this limitation, dynamic airspace allocation systems can be introduced to designate specific corridors for AAM operations \citep{abdellaoui2023building,shrestha2021survey}. These corridors can be managed through time-based scheduling mechanisms that allocate flight windows based on demand, safety conditions, and traffic density. Such flexible access structures optimize airspace utilization while maintaining safety standards \citep{abdellaoui2023building}. This approach ensures that regulatory control is preserved while operational adaptability is significantly enhanced, thereby reducing perceptions of excessive restriction.

\vspace{0.2cm}

\noindent \textbf{iv. Transparent and Standardized Regulatory Governance Systems}: Another important dimension of regulatory concern relates to inconsistency and lack of transparency in enforcement and interpretation across different regulatory bodies. Uncertainty regarding differences between federal and local regulations can further complicate compliance processes. To address this, regulatory information should be consolidated into unified and publicly accessible governance portals that provide standardized guidance. In addition, maintaining regularly updated regulatory documentation, interpretation manuals, and clearly defined enforcement criteria can reduce ambiguity and improve predictability in compliance expectations \citep{cartile2025digital,naevestad2021can}. Establishing consistent communication channels between regulators and operators further enhances trust by enabling clarification of compliance requirements and reducing uncertainty in regulatory interpretation.

\vspace{0.2cm}

\noindent \textbf{v. Accessible Education and Streamlined Certification Pathways}: Improving accessibility to regulatory education and certification processes can lower entry barriers for new AAM participants. Complex licensing structures can be simplified through modular certification pathways that allow incremental progression based on competency levels. Simulation-based training environments can further support learners in understanding regulatory requirements in practical scenarios \citep{wikander2016multi,zhang2025evaluating,jalandharachari2026integrating}. In parallel, awareness programs that clearly communicate operational rules, safety expectations, and legal responsibilities can reduce confusion and improve compliance readiness. By making regulatory education more accessible and structured, individuals are better equipped to navigate certification processes, thereby improving overall system adoption and regulatory alignment.

\subsubsection{Strategies to Address Concerns Related to Technical Performance of Drones}

Technical, hardware, and flight performance-related concerns emerge as the third most frequently discussed cluster of concern in our analysis. In negative, neutral, and exploratory discussions, users frequently question whether current technological capabilities are sufficient to support sustained and commercially viable AAM operations under real-world conditions. Although these concerns do not reflect opposition to AAM itself, they highlight perceived engineering limitations that could slow adoption and reduce confidence in operational readiness. To address these challenges and strengthen positive adoption, five distinct technical improvement strategies can be implemented.

\vspace{0.2cm}

\noindent \textbf{i. Enhancing energy and propulsion performance}: A central technical concern relates to energy endurance and propulsion efficiency, particularly the limitations in current battery systems for sustained flight operations. Current drone batteries, including lithium-ion and lithium-polymer systems, typically achieve energy densities of approximately 150--300 Wh/kg at the cell level \citep{batterytips2025drones}, which directly limits flight endurance to short durations (typically tens of minutes under operational load) and remains insufficient for long-endurance and high-payload AAM missions \citep{zhang2024overall}. As a result, improving battery energy density would increase flight time and operational range, enabling longer missions and broader deployment scenarios. Addressing this limitation requires continued development of next-generation energy storage technologies, including high-density batteries, hybrid-electric propulsion systems, and lightweight energy solutions optimized for AAM use cases \citep{kiesewetter2023holistic,schuchardt2026scientific}. 


\vspace{0.2cm}

\noindent \textbf{ii. Integrated airframe and system architecture optimization}: Another major concern involves trade-offs between payload capacity, energy consumption, and structural weight, meaning that improving one aspect often negatively affects the others due to physical design limits in aircraft systems. For example, increasing payload adds weight, which increases energy consumption and reduces flight endurance, while reducing structural weight can limit strength or safety margins. Improving these aspects requires advances in airframe design, including lightweight composite materials, aerodynamic optimization, and modular structural configurations adaptable to different mission requirements \citep{siengchin2023review,johnson2026impact}. Rather than treating propulsion, energy storage, and structure as separate subsystems, integrated system-level co-design allows these components to be optimized together. This reduces inefficiencies caused by mismatched subsystem design and improves overall performance. As a result, this unified design approach can enhance both endurance and payload capacity simultaneously.

\vspace{0.2cm}

\noindent \textbf{iii. Reliability assurance through certification and pre-deployment validation}: Ensuring system reliability is a fundamental requirement in AAM, as current systems are still affected by firmware instability, hardware malfunctions, and electronic component failures, as evidenced by discussions in social media posts within our AAM dataset. These concerns highlight the need for strong pre-deployment validation mechanisms that ensure safety and performance before operational use. To address this, rigorous testing protocols, stress-testing procedures, and aviation-grade certification standards should be implemented for AAM components. These processes verify that both hardware and software meet defined reliability thresholds prior to deployment, improving trust in operational safety and technical dependability \citep{schuchardt2026scientific,yoo2022risk}.

\vspace{0.2cm}

\noindent \textbf{iv. Incremental deployment and post-deployment operational learning}: Technical confidence can be strengthened through phased deployment rather than immediate large-scale commercialization. Initial operations in controlled environments, short-range missions, and low-payload scenarios allow systems to be tested under reduced risk conditions while performance is gradually validated in real-world settings \citep{nilsson2024advanced}. This staged approach enables systematic evaluation of system performance across increasing levels of operational complexity before full-scale deployment \citep{mankins2009technology}. Once deployed, continuous operational feedback becomes essential for long-term improvement. Data from system logs, maintenance records, and pilot feedback can be used to identify failures and performance limitations, enabling iterative refinement of both hardware and software components \citep{hu2025methodology}.

\subsubsection{Strategies for Reducing Military and Geopolitical Concerns for AAM Adoption}

Military, geopolitical, and defense-related concerns represent the fourth most discussed cluster of concern and are primarily driven by perceptions of drones in military applications, including swarm operations, autonomous weapon systems, and their use in conflict and surveillance scenarios, as well as broader concerns about geopolitical tensions and global security implications. To address these perceptions and support civilian adoption, two major strategies can be implemented.

\vspace{0.2cm}

\noindent \textbf{i. Reinforcing civilian identity through structural separation and emphasizing humanitarian applications}: A key strategy is to strengthen the perception of AAM as a civilian transportation technology by combining institutional separation with visible non-military use cases. This involves establishing dedicated civilian certification categories, airworthiness standards, and regulatory frameworks that clearly define AAM as a system for passenger transport, logistics, emergency response, and urban mobility rather than defense applications \citep{clothier2015risk,aydin2019public,pongsakornsathien2025advances,pons2022understanding}. In parallel, this civilian identity is reinforced by emphasizing real-world humanitarian and societal applications of AAM technologies. Use cases such as medical supply delivery, disaster response, emergency evacuation support, and connectivity to remote areas demonstrate tangible public value and help reposition AAM as critical infrastructure rather than a security threat \citep{sigari2021medical}. Evidence from operational deployments in healthcare logistics and emergency management further strengthens this framing by providing concrete proof of non-military utility. Together, structural separation and application visibility reduce ambiguity and weaken associations between AAM and military drone systems.

\vspace{0.2cm}

\noindent \textbf{ii. Strengthening transparent governance and advancing international coordination}: A second strategy focuses on improving trust through governance at both institutional and global levels. At the institutional level, public confidence can be enhanced by ensuring that AAM systems operate under clearly defined safety standards, audit mechanisms, and accountability frameworks \citep{garcia2026governance,dolata2023moving}. Transparent disclosure of operational boundaries, system design principles, and safety protocols reduces uncertainty and addresses concerns related to misuse or covert surveillance. Independent oversight structures and publicly accessible safety reporting further reinforce regulatory accountability and trust. At the international level, reducing geopolitical concerns requires harmonized global frameworks for AAM operations, including airspace management, communication protocols, and safety certification standards \citep{schuchardt2026scientific,barr2026cross}. Coordinated regulatory development across aviation authorities reduces fragmentation and mitigates perceptions of technological competition or strategic dominance. This dual-layer governance approach positions AAM as a globally coordinated civil aviation innovation governed by transparent and consistent rules.

\subsubsection{Strategies to Address Safety and Operational Risks in AAM}

According to the Cross-Sentiment Topic Analysis, safety-related concerns about drones and emerging flying vehicles correspond to the fifth cluster of concern and represent the most dominant topic within the negative sentiment class alone. These concerns arise primarily from fears of mid-air collisions, system malfunctions, crashes in populated areas, and a perceived lack of strong safety assurances for routine aerial operations. Although these issues reflect increased sensitivity among users on public platforms rather than opposition to the technology itself, they have a substantial impact on public trust and willingness to support AAM deployment. To address these challenges, five distinct and complementary strategies can be implemented.

\vspace{0.2cm}

\noindent \textbf{i. Safety-by-design through redundancy and fail-operational architectures}: A primary strategy is to embed safety directly into the vehicle architecture by ensuring redundancy across critical systems. AAM vehicles are designed with multiple navigation units, backup power systems, and automated emergency landing capabilities that maintain control even under partial system failure conditions \citep{xu2025systematic}. Distributed Electric Propulsion (DEP), as demonstrated in systems such as Joby Aviation, eliminates single points of failure by allowing remaining motors to dynamically compensate and maintain stable flight \citep{swartz2020jobyunicorn}. This approach focuses on preventing catastrophic failure through inherent system design.

\vspace{0.2cm}

\noindent \textbf{ii. Pre-deployment certification and safety validation}: A second strategy focuses on verifying system safety before operational deployment. This includes structured certification processes involving simulation-based stress testing, real-world flight trials, and phased approval stages aligned with aviation-grade standards \citep{altun2023development}. In this context, Joby Aviation was among the first AAM developers to receive final airworthiness criteria approval from the FAA, representing an important regulatory milestone and contributing to the establishment of early safety and certification benchmarks for the emerging industry \citep{Weitering2024JobyFAA}. Complementing such regulatory advancements, recent research efforts such as NASA’s collision risk modeling for eVTOL operations under Required Navigation Performance (RNP) 0.3 conditions indicate a minimum safe separation threshold of approximately 1.1–1.2 nautical miles, assuming lateral navigation accuracy within ±0.3 NM for 95\% of operations 
\citep{bickmeier2025collision}. These findings demonstrate how quantitative safety thresholds can be derived from high-fidelity simulation environments to inform certification and operational requirements. Accordingly, clearly defined safety thresholds and performance benchmarks ensure that systems meet minimum reliability requirements before entering service, including collision avoidance constraints, system redundancy requirements, and navigation accuracy standards derived from simulation-based risk assessments and experimental validation studies. However, more empirical studies of this type are still needed to systematically define robust safety thresholds and operational requirements for large-scale AAM deployment, and this strategy emphasizes validation prior to deployment rather than in-operation mitigation.

\vspace{0.2cm}

\noindent \textbf{iii. Real-time system monitoring and predictive maintenance}: Operational safety can be enhanced through continuous monitoring of aircraft health during service. Advanced sensor networks and AI-based diagnostic systems enable real-time detection of anomalies in propulsion, navigation, and control systems \citep{fu2023prognostic}. Predictive maintenance frameworks allow early intervention before faults develop into safety-critical failures, reducing unexpected malfunctions and improving operational stability \citep{fu2023prognostic}. This strategy focuses on preventing failures during active operation.

\vspace{0.2cm}

\noindent \textbf{iv. Airspace risk management through structured operational design}: Safety risks can also be reduced by controlling the operational environment of AAM systems. This involves establishing geofenced flight corridors, restricting operations in densely populated or sensitive areas, and implementing structured low-altitude airspace management systems \citep{pongsakornsathien2025advances}. Dynamic airspace control further enhances safety by adjusting flight permissions based on real-time traffic density and environmental conditions. This strategy reduces external exposure to risk by limiting where and when operations occur.

\vspace{0.2cm}

\noindent \textbf{v. Transparency and public communication of safety performance}: Finally, public confidence is supported through transparent reporting of safety performance and operational outcomes. Publishing certification results, incident data, and operational safety metrics in accessible formats helps reduce uncertainty and misinformation \citep{lindhout2022transparency}. Continuous disclosure of performance improvements over time supports public understanding of system reliability and reinforces accountability in AAM operations.

\subsubsection{Strategies to Address Noise and Perceived Disturbance in AAM}

Noise and perceived disturbance represent the least discussed but still important cluster of concern in the AAM discourse, reflecting frustration with the auditory impact of drones and aerial vehicles during operation. These concerns are primarily driven by perceptions of constant or high-frequency sound, disruption of daily activities, and reduced quality of life in residential, urban, and suburban environments. Although not as frequently discussed as other issues, noise pollution can influence community acceptance and the long-term social sustainability of AAM operations. To address these concerns, several distinct and complementary strategies can be implemented.

\vspace{0.2cm}

\noindent \textbf{i. Noise reduction through vehicle-level acoustic and propulsion design}: A primary strategy is to reduce noise at its source through improvements in aircraft design and propulsion systems. This includes quieter electric motors, optimized rotor blade geometry, distributed propulsion configurations, and aerodynamic shaping that reduces turbulence and tonal noise generation \citep{schade2025psychoacoustic,tripaldi2025emerging}. Industry examples indicate progress in lowering acoustic signatures; for instance, Joby Aviation reports cruise noise levels of approximately 45.2 dBA at 1,640 ft altitude, comparable to a quiet office environment \citep{joby2022lownoise}. Similarly, Archer Aviation reports significantly reduced acoustic output relative to conventional helicopters, with cruise conditions around ~45 dBA. This strategy focuses on minimizing noise production at the source through engineering design.

\vspace{0.2cm}

\noindent \textbf{ii. Exposure management through operational control and urban integration}: A second strategy focuses on reducing human exposure to noise by jointly optimizing operational patterns and urban airspace planning. This includes adjusting flight paths to avoid densely populated areas, regulating cruise altitudes to reduce ground-level sound exposure, and establishing dedicated aerial corridors designed to minimize residential disturbance \citep{bauranov2021designing}. Time-based restrictions, such as limiting operations during night hours or peak residential activity periods, further reduce disturbance \citep{bauranov2021designing,gao2024noise}. In parallel, urban planning considerations ensure that vertiports, landing zones, and flight corridors are located in areas that minimize long-term population exposure to noise \citep{gao2024noise}. This combined approach integrates operational scheduling with infrastructure placement to reduce cumulative noise impact on communities.

\vspace{0.2cm}

\noindent \textbf{iii. Continuous noise monitoring and adaptive control systems}: Operational noise can also be managed dynamically through real-time monitoring systems. Acoustic sensors deployed in operational areas can measure sound levels continuously and detect exceedances of predefined thresholds \citep{giladi2020real,liu2025iot}. Based on this data, flight paths or operational intensity can be adjusted in real time to reduce disturbance. This creates a feedback-driven system for maintaining compliance with acceptable noise limits during operations.

\vspace{0.2cm}

\noindent \textbf{iv. Public engagement and transparency in noise mitigation}: Acceptance of AAM depends on the transparent communication of noise impacts and mitigation outcomes. Providing accessible data from testing, pilot programs, and operational deployments helps demonstrate actual noise levels and improvements over time \citep{eissfeldt2022public,smith2022public,gao2024noise}. Public engagement initiatives that explain mitigation strategies and performance results further support trust and reduce perception gaps between expected and actual noise impact.

\section{Conclusion and Future Work}

This study addresses the critical need to systematically understand public sentiment toward AAM at scale, as traditional survey-based approaches are insufficient to capture the breadth, dynamism, and evolving nature of public discourse surrounding emerging aviation systems. To this end, we developed an AI-driven analytical framework applied to 306,009 user-generated texts from Reddit and Quora, integrating sentiment classification, topic modeling, temporal trend analysis, and cross-sentiment comparison. A carefully annotated subset of 5,000 texts, labeled using a hybrid approach involving two human annotators and an OpenAI-based annotator, ensured labeling reliability and supported model evaluation and fine-tuning. Comparative experiments across seven sentiment analysis models demonstrate that transformer-based architectures outperform traditional machine learning and lexicon-based approaches, with ModernBERT achieving the highest classification accuracy.


Building on the sentiment labels, topic modeling identified 20 latent topics distributed across four sentiment classes: six in Positive, six in Negative, four in Neutral, and four in Others, enabling a fine-grained analysis of AAM discourse. Within the Positive class, \emph{Drone Hardware and Flight Control} (21.55\% of the dataset) and \emph{Personal Experiences of Using Drones} (21.46\%) are the dominant topics, indicating that positive sentiment is driven by usability and direct engagement. In contrast, the Negative class is primarily shaped by \emph{Regulation and Compliance} (21.05\%) and \emph{Safety Concerns} (20.90\%), reflecting governance-related constraints and perceived operational risks. Similarly, the Neutral class is dominated by \emph{Drone Industry, Workforce, and Production} (31.45\%), highlighting informational and descriptive discourse, while the Others class is led by \emph{Emerging Drone Applications} (28.89\%), emphasizing exploratory and future-oriented discussions. 
Furthermore, cross-sentiment analysis consolidates these findings into six major concern clusters, where \emph{Workforce, Skill Development, and Career Pathways} emerges as the most dominant cluster (25.29\%), followed by \emph{Regulation and Compliance} (24.64\%) and \emph{Technical Performance} (20.99\%), while \emph{Military, Geopolitics, and Defense} accounts for 14.58\%, and \emph{Safety and Operational Risks} (8.51\%) along with \emph{Noise and Disturbance} (5.98\%) form smaller yet consistently negative clusters, both entirely dominated by negative sentiment.

While this study provides a large-scale analysis of public sentiment toward AAM, several directions remain for future research. First, the current framework can be extended by incorporating additional data sources, such as news articles, policy documents, and aviation-related forums, to capture a broader and more comprehensive view of public discourse. Second, future work can explore multimodal sentiment analysis by integrating textual data with other modalities, such as images, videos, and audio content, which may provide deeper insights into how AAM is perceived across different communication formats. Another important direction for future work is the development of an interactive public interface built on the proposed analytical framework. Leveraging the curated dataset, fine-tuned sentiment models, and generative AI capabilities, such an interface could enable real-time querying and exploration of public sentiment insights. The system would dynamically generate responses to user queries by synthesizing model outputs, providing concise summaries, and highlighting key areas of public concern. This direction would bridge the gap between advanced AI analytics and practical usability, transforming static analysis into an interactive decision-support tool for both public and policy-level stakeholders. Finally, future work can further investigate the effectiveness of the proposed strategies by examining how public sentiment evolves following the implementation of AAM-related policies. In particular, causal relationships between policy events (e.g., FAA announcements, prototype demonstrations, and infrastructure deployments) and shifts in public sentiment can be analyzed to better understand the drivers of perception change over time. Such analysis would help determine whether interventions designed to address key concerns actually translate into measurable improvements in public acceptance.

\section*{Acknowledgment}

The authors would like to acknowledge Zaynah Wahab for her contribution during the early phase of this research on AAM public sentiment analysis.

\section*{Data Statement}

All collected data were publicly available and did not include any personally identifiable information. The dataset was used solely for research purposes in accordance with standard ethical guidelines for computational text analysis.



\begin{spacing}{1}
\bibliography{cas-refs}
\end{spacing}

\end{document}